\documentclass[journal,twoside,web]{ieeecolor}
\usepackage{etoolbox}
\makeatletter
\@ifundefined{color@begingroup}%
{\let\color@begingroup\relax
\let\color@endgroup\relax}{}%
\def\fix@ieeecolor@hbox#1{%
\hbox{\color@begingroup#1\color@endgroup}}
\patchcmd\@makecaption{\hbox}{\fix@ieeecolor@hbox}{}{\FAILED}
\patchcmd\@makecaption{\hbox}{\fix@ieeecolor@hbox}{}{\FAILED}
\usepackage{tmi}
\usepackage{cite}
\usepackage{amsmath,amssymb,amsfonts}
\usepackage{algorithm}
\usepackage{algpseudocode}
\usepackage{graphicx}
\usepackage{textcomp}
\usepackage[utf8]{inputenc}
\usepackage{tabularx}
\usepackage{multirow}
\usepackage{booktabs}
\usepackage{colortbl}
\usepackage[colorlinks,linkcolor=blue]{hyperref}
\def\BibTeX{{\rm B\kern-.05em{\sc i\kern-.025em b}\kern-.08em
    T\kern-.1667em\lower.7ex\hbox{E}\kern-.125emX}}
\begin{document}
\title{Toward Reliable AR-Guided Surgical Navigation: Interactive Deformation Modeling with Data-Driven Biomechanics and Prompts}
\author{Zheng Han, Jun Zhou, Jialun Pei, \IEEEmembership{Member, IEEE}, Jing Qin, \IEEEmembership{Senior Member, IEEE}, Yingfang Fan, \\ and Qi Dou, \IEEEmembership{Senior Member, IEEE}
\thanks{This work was supported in part by the Hong Kong Innovation and Technology Fund (Project No. GHP/167/22SZ), in part by the Research Grants Council of Hong Kong Special Administrative Region, China (Projects No. N\_CUHK410/23 and No. T45-401/22-N), in part by the National Natural Science Foundation of China (Project No. 62322318), in part by the General Research Fund of Hong Kong Research Grants Council (Project No. 15218521), and in part by the Shenzhen Science and Technology Program (Project No. SGDX20230116092200001). We sincerely thank Dr. Kai Wang from the Department of General Surgery at Nanfang Hospital in Guangzhou, China, for his clinical support. \emph{Corresponding authors: Yingfang Fan and Qi Dou}}
\thanks{Zheng Han, Jialun Pei, and Qi Dou are with the Department of Computer Science and Engineering, The Chinese University of Hong Kong, HKSAR, China. (e-mail: zheng-han@link.cuhk.edu.hk).}
\thanks{Jun Zhou and Jing Qin are with the Center of Smart Health, School of Nursing, The Hong Kong Polytechnic University, HKSAR, China.}
\thanks{Yingfang Fan is with the Department of Hepatobiliary Surgery, The Third Affiliated Hospital of Southern Medical University, Guangzhou, China.}
\thanks{DOI: 10.1109/TMI.2025.3577759}
}
\maketitle

\begin{abstract}
In augmented reality (AR)-guided surgical navigation, preoperative organ models are superimposed onto the patient’s intraoperative anatomy to visualize critical structures such as vessels and tumors. Accurate deformation modeling is essential to maintain the reliability of AR overlays by ensuring alignment between preoperative models and the dynamically changing anatomy. Although the finite element method (FEM) offers physically plausible modeling, its high computational cost limits intraoperative applicability. Moreover, existing algorithms often fail to handle large anatomical changes, such as those induced by pneumoperitoneum or ligament dissection, leading to inaccurate anatomical correspondences and compromised AR guidance.
To address these challenges, we propose a data-driven biomechanics algorithm that preserves FEM-level accuracy while improving computational efficiency. In addition, we introduce a novel human-in-the-loop mechanism into the deformation modeling process. This enables surgeons to interactively provide prompts to correct anatomical misalignments, thereby incorporating clinical expertise and allowing the model to adapt dynamically to complex surgical scenarios.
Experiments on a publicly available dataset demonstrate that our algorithm achieves a mean target registration error of 3.42~mm. Incorporating surgeon prompts through the interactive framework further reduces the error to 2.78~mm, surpassing state-of-the-art methods in volumetric accuracy.
These results highlight the ability of our framework to deliver efficient and accurate deformation modeling while enhancing surgeon-algorithm collaboration, paving the way for safer and more reliable computer-assisted surgeries.

\end{abstract}

\vspace{1.0em}

\begin{IEEEkeywords}
Augmented reality, human-in-the-loop, non-rigid registration, surgical navigation.
\end{IEEEkeywords}

\section{Introduction}
\label{sec:introduction}
\begin{figure}[!t]
\centerline{\includegraphics[width=1.0\columnwidth]{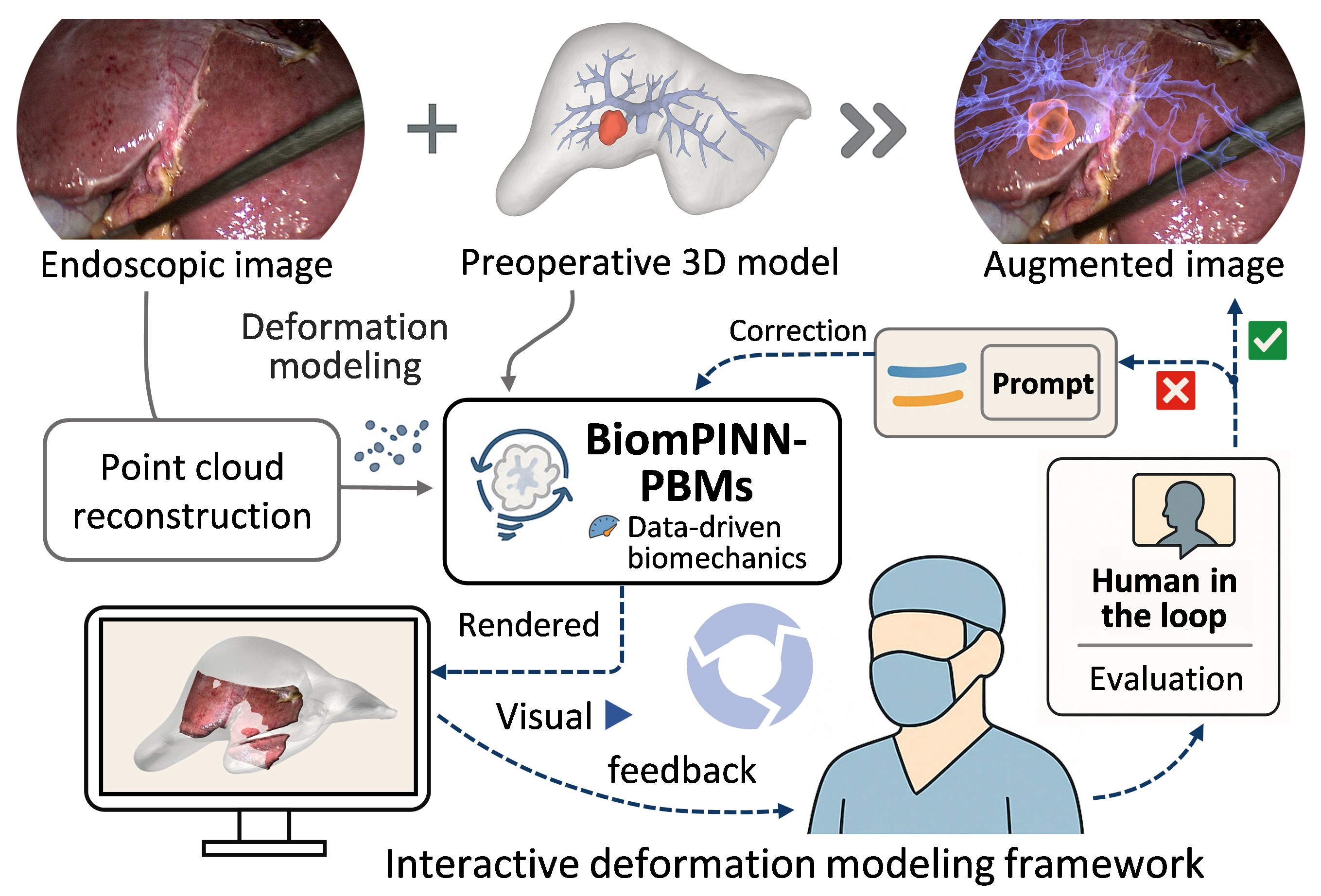}}
\caption{Interactive deformation modeling for AR-guided surgical navigation. The proposed data-driven biomechanics algorithm (BiomPINN-PBMs) deforms the preoperative 3D model based on intraoperative imaging and presents the result for surgeon evaluation. This human-in-the-loop process allows the surgeon to either accept the deformation for AR overlay or provide corrective prompts to further refine the model. A video demonstration of the interactive process is available at: \href{https://github.com/med-air/Interactive-registration}{https://github.com/med-air/Interactive-registration}.}
\label{fig:teaser}
\vspace{-10pt}
\end{figure}

\IEEEPARstart{A}{ugmented} Reality (AR) has the potential to transform surgical procedures by intuitively visualizing critical anatomical structures within the surgical field. This is achieved by superimposing virtual organ models—typically reconstructed from preoperative computed tomography (CT) or magnetic resonance imaging (MRI) scans—onto the patient’s intraoperative anatomy \cite{birlo2022utility}. As illustrated in Fig.~\ref{fig:teaser}(top), AR overlays vascular structures and tumor locations onto live endoscopic views, thereby providing surgeons with spatially contextualized guidance during complex procedures such as tumor resections or vessel preservation. Such enhanced visualization is especially critical in minimally invasive surgery, where the field of view is narrow and tactile feedback is limited. Accurate alignment between virtual models and intraoperative anatomy enables the identification of structures that are occluded or difficult to discern from endoscopic views alone.

\includegraphics[width=0pt,height=0pt]{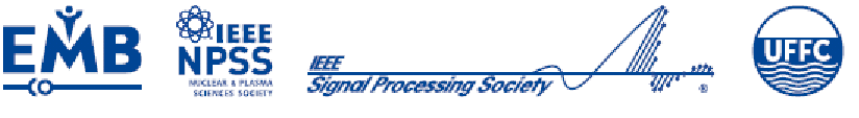}

However, preoperative models represent static anatomy captured days or weeks prior to surgery and therefore fail to reflect the dynamic intraoperative environment. Factors such as pneumoperitoneum pressure, instrument interaction, and ligament dissection can induce substantial deformation of soft tissues, leading to spatial misalignment between the virtual model and actual anatomy \cite{luo2020augmented}. Without addressing these deformations, AR overlays may become inaccurate or misleading, potentially compromising surgical safety and outcomes.

Patient-specific biomechanical models (PBMs) have been widely adopted to compensate for intraoperative deformations, offering physically plausible representations of soft tissue viscoelasticity \cite{plantefeve2016patient}. Grounded in continuum mechanics, PBMs typically employ finite element (FE) methods to simulate organ deformation under prescribed material properties, boundary conditions, and loading scenarios \cite{mass2012febio}. However, acquiring comprehensive input data for FE simulations in the surgical setting is often infeasible. Key parameters—such as contact forces and torques—are difficult to measure due to their invasive nature and the lack of integrated sensors in surgical instruments \cite{pfeiffer2020non}. As a result, intraoperative information is usually limited to visual observations, such as laparoscopic images or reconstructed organ surfaces, which only capture partial and superficial aspects of deformation. Inferring full volumetric deformation from such sparse and incomplete surface cues remains a fundamental challenge to achieving anatomically accurate AR navigation.

\begin{figure}[!t]
\centerline{\includegraphics[width=0.95\columnwidth]{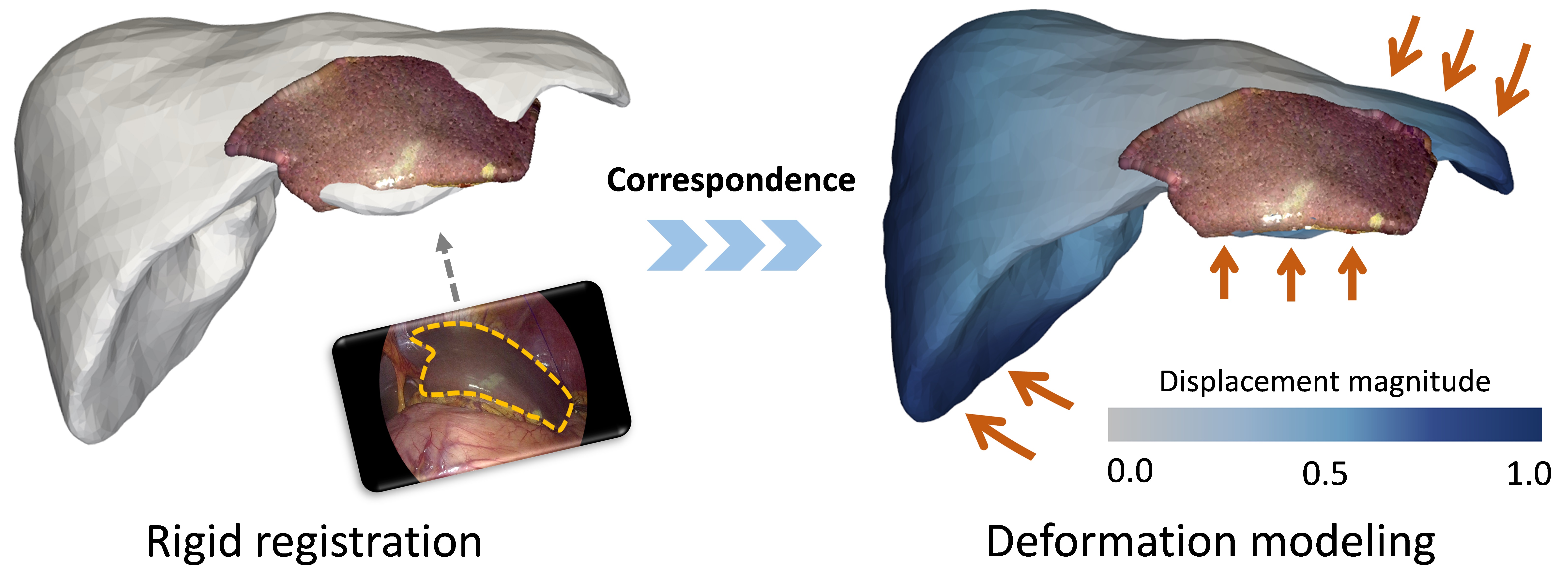}}
\caption{Illustration of the surface-based deformation modeling process. After rigid registration, sparse correspondences with the intraoperative surface are established (left). These local displacements are biomechanically propagated to estimate the full volumetric deformation field, including unobserved regions (right).}
\label{fig:modeling_process}
\vspace{-10pt}
\end{figure}

To address these practical limitations, image-driven surface-based PBMs have been developed \cite{cash2005compensating, tagliabue2021data, min2023non, khallaghi2015biomechanically, mestdagh2022optimal, mendizabal2023intraoperative}. These methods infer organ deformation by leveraging sparse surface displacements derived from geometric correspondences between the preoperative model and intraoperative observations, as illustrated in Fig.~\ref{fig:modeling_process}. The observed displacements are imposed as Dirichlet boundary conditions to drive the surface alignment \cite{tagliabue2021data}, while FE-based regularizers are introduced to ensure biomechanical plausibility beyond observed regions. A critical component of this formulation is the choice of regularization weight, which balances geometric fidelity against physical plausibility. Overly stiff regularization may hinder surface alignment, whereas overly flexible settings can lead to non-physical deformations. To achieve an optimal balance, iterative optimization schemes are employed, gradually tuning the regularization weight to minimize alignment error while maintaining deformation coherence \cite{khallaghi2015biomechanically}. However, such iterative tuning is computationally expensive, as each iteration requires solving a large-scale linear system governed by the stiffness matrix. Furthermore, the inherently sequential nature of these solvers limits their ability to leverage parallel computation on modern hardware architectures, further constraining their applicability in time-sensitive surgical scenarios \cite{naumov2012parallel}.

In parallel, deep learning–based models have been explored as alternatives to PBMs, aiming to approximate biomechanical deformation with lower computational costs \cite{zhu2022real}. Early approaches \cite{pfeiffer2019learning, pfeiffer2020non} discretized organs into voxelized 3D grids and employed convolutional neural networks (e.g., U-Net) to regress displacement fields. While effective in capturing global deformation patterns, these voxel-based approaches are constrained by limited resolution and voxelization artifacts. To overcome these issues, more recent works have adopted flexible geometric representations, such as point clouds \cite{azampour2024anatomy} and implicit neural fields \cite{jia2021improving, henrich2024tracking}, which better preserve fine anatomical details and enable richer deformation modeling. However, these representations often lack explicit spatial coupling among neighboring points, making it difficult to propagate local displacements into globally consistent volumetric deformation. Consequently, predicted strain fields frequently exhibit physical inconsistencies, including non-smoothness and discontinuities \cite{heiselman2024image}. Another major challenge is data availability. Ground truth volumetric deformations are rarely obtainable in surgical settings, as acquiring them would require repeated intraoperative CT or MRI scans—procedures that are both ethically and logistically impractical. As a result, most data-driven models are trained on synthetic datasets generated through forward FE simulations. This offline training–testing paradigm introduces a domain shift: models trained on simulated anatomies and deformation patterns often fail to generalize to real, patient-specific conditions. Thus, despite their computational efficiency, current data-driven approaches remain constrained in clinical deployment by distribution shifts between training and intraoperative data and by insufficient guarantees of physical plausibility.

Additionally, complex surgical scenarios involving large anatomical deformations pose inherent challenges for surface-based deformation modeling methods, including both physics-based and data-driven approaches. These methods rely on establishing accurate correspondences between the preoperative model and intraoperative anatomy—a critical prerequisite for driving deformation using sparse surface displacements. However, under conditions of severe morphological change, such correspondences often become unreliable. For example, in liver surgery, resection of the falciform ligament can induce structural alterations, such as pronounced separation between the left and right lobes, that are not reflected in the preoperative model. This anatomical discrepancy, compounded by the lack of distinctive geometric features on the organ surface, further hampers the establishment of reliable correspondences. As a result, such challenging scenarios often exceed the capabilities of fully automated pipelines, leading to inaccurate alignments and compromising the reliability of AR-guided navigation. These limitations underscore the urgent need for an adaptive framework that incorporates human expertise to correct deformation results when automated methods fall short \cite{han2024review}.

In summary, we identify three key challenges in intraoperative deformation modeling: (1) the high computational burden associated with biomechanical models; (2) the lack of guaranteed physical plausibility in data-driven predictions, compounded by domain shift when applying synthetic-data-trained models to real surgical settings; and (3) the limited effectiveness of existing methods under scenarios involving large anatomical deformations.

In response to these challenges, we present the following contributions:

\begin{itemize}
     \item \textbf{Biomechanically-Constrained Neural Network:}
     We introduce \textit{BiomPINN}, a \textbf{P}hysics-\textbf{I}ntegrated \textbf{N}eural \textbf{N}etwork that incorporates \textbf{Biom}echanical priors to enforce physical laws and biomechanical plausibility. BiomPINN adopts a \textit{per-instance optimization paradigm}, eliminating the need for offline training and avoiding distribution shift between synthetic and intraoperative data.

    \item \textbf{BiomPINN-PBMs Integration:} We present a hybrid approach that integrates BiomPINN with PBMs, preserving FE–level deformation accuracy while significantly reducing computational cost through one-pass optimization.
    
    \item \textbf{Interactive Deformation Modeling Framework:} We develop an interactive, human-in-the-loop framework tailored for AR-guided surgery. When automated predictions are unsatisfactory, surgeons can provide intuitive corrective prompts, which are incorporated into the modeling process to dynamically adapt to complex intraoperative scenarios.
    
    \item \textbf{Superior Performance:} Our approach demonstrates superior volumetric deformation accuracy compared to state-of-the-art methods, validated on a publicly available phantom dataset.

\end{itemize}

\section{Preliminary Information}

Given two organ surfaces—a source point cloud \( \mathcal{X} = \{ \mathbf{x}_i \in \mathbb{R}^3 \mid i = 1, \dots, n \} \), representing the complete geometric structure of the organ surface reconstructed from preoperative CT, and a target point cloud \( \mathcal{Y} = \{ \mathbf{y}_i \in \mathbb{R}^3 \mid i = 1, \dots, m \} \), representing an incomplete organ surface reconstructed from intraoperative imaging—the objective is to determine a deformation field \( \mathcal{U} = \{ \mathbf{u}_i \in \mathbb{R}^3 \mid i = 1, \dots, n \} \) that warps the preoperative model \( \mathcal{X} \) to align with the intraoperative observations \( \mathcal{Y} \).

To ensure that the deformation of \( \mathcal{X} \) follows physically plausible behavior, a biomechanical model is introduced as a global geometric constraint to enforce volumetric consistency.

\subsection{Biomechanical Model Representation}
The biomechanical model is defined as a volumetric tetrahedral mesh \( \Omega \), constructed based on the surface node set \( \mathcal{X} \), with internal nodes automatically generated during the tetrahedralization process. The mesh \( \Omega \) consists of \( n_v \) nodes, denoted as \(\mathcal{X}_{\Omega} = \{ \mathbf{x}_{\Omega,i} \in \mathbb{R}^3 \mid i = 1, \dots, n_v \} \in \mathbb{R}^{3n_v}. \) The point set \( \mathcal{X} \) is a subset of the nodes in \( \Omega \) and forms a closed surface mesh, denoted by \( \partial \Omega \).

When surface tractions are applied on the boundary \( \partial \Omega \), the resulting displacements within the volumetric domain \( \Omega \), denoted as \(\mathcal{U}_{\Omega} = \{ \mathbf{u}_{\Omega,i} \in \mathbb{R}^3 \mid i = 1, \dots, n_v \},\) can be obtained by solving the static equilibrium equation:
\begin{equation}
\mathbf{K} \mathbf{u}_{\Omega} = \mathbf{f}, \label{eq:static_equilibrium}
\end{equation}
where \( \mathbf{u}_{\Omega} \in \mathbb{R}^{3n_v} \) is the global displacement vector obtained by stacking all nodal displacements in \( \mathcal{U}_{\Omega} \), and \( \mathbf{K} \in \mathbb{R}^{3n_v \times 3n_v} \) is the global stiffness matrix, which encodes the organ’s geometry and material properties as defined by the FE model. The force vector \( \mathbf{f} \) represents the nodal forces assembled from the applied surface tractions on \( \partial \Omega \).

\begin{figure}[!t]
\centerline{\includegraphics[width=0.95\columnwidth]{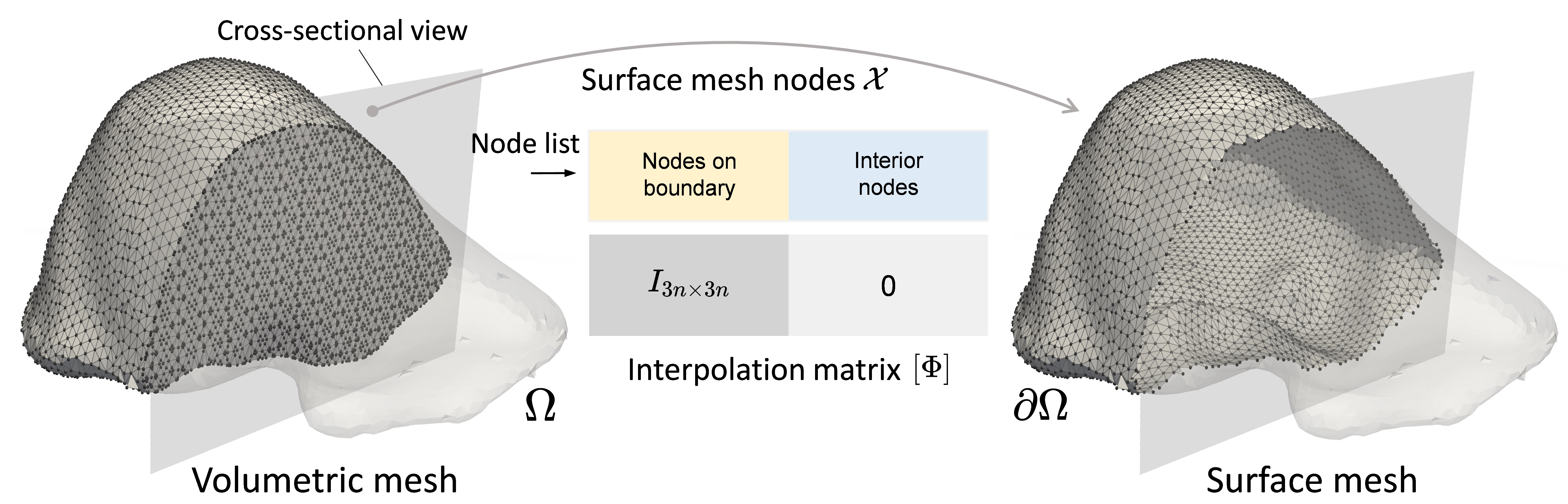}}
\caption{Illustration of the interpolation matrix \( \Phi \) that maps volumetric displacements to boundary displacements. The tetrahedral mesh \( \Omega \) is partitioned into boundary nodes \( \mathcal{X} \subset \partial \Omega \) and interior nodes, yielding a block matrix structure. A cross-sectional view reveals the internal mesh structure, while the surface mesh \( \partial \Omega \) retains only the boundary points.}
\label{fig:interpolation_matrix}
\vspace{-10pt}
\end{figure}

To relate the volumetric displacements to the surface boundary displacements, an interpolation matrix \( \Phi \in \mathbb{R}^{3n \times 3n_v} \) is introduced:
\begin{equation}
\mathbf{u} = \Phi\, \mathbf{u}_{\Omega}, \label{eq:interpolation}
\end{equation}
where \( \mathbf{u} \in \mathbb{R}^{3n} \) is the vectorized displacement of the surface nodes, obtained by stacking the deformation field \( \mathcal{U} \). Following the node ordering strategy in~\cite{khallaghi2015biomechanically}, the nodes in \( \mathcal{X}_{\Omega} \) that exclusively belong to the surface \( \partial \Omega \) are appended to the end of the node list. Consequently, the interpolation matrix \( \Phi \) takes the following block form:
\begin{equation}
\Phi = \begin{bmatrix}
I_{3n \times 3n} & 0
\end{bmatrix}.
\label{eq:phi_structure}
\end{equation}
Here, the identity block extracts the displacements of surface nodes (assumed to be ordered first), while the zero block corresponds to the remaining internal nodes. This matrix provides a linear mapping from the full volumetric displacement field to surface displacements, enabling the use of the biomechanical model as a constraint in surface-based deformation modeling (see Fig.~\ref{fig:interpolation_matrix} for illustration).

\subsection{Stiffness Matrix}
The biomechanical model's geometry and material properties are encoded in the stiffness matrix \( \mathbf{K} \) of the FE model. As shown in Eq.~\eqref{eq:static_equilibrium}, the global stiffness matrix \( \mathbf{K} \) establishes the relationship between the nodal displacement vector \( \mathbf{u}_{\Omega} \) and the force vector \( \mathbf{f} \). For a linearly elastic tissue model~\cite{bonet1997nonlinear}, the mechanical behavior is characterized by two material constants: Young's modulus (\( E \)) and Poisson's ratio (\( \nu \)), which define the constitutive relationship between stress and strain.

In practice, the global stiffness matrix \( \mathbf{K} \) is assembled from the local contributions of all elements in the tetrahedral mesh. For an individual element \( e \), the local stiffness matrix \( \mathbf{K}_e \) is given by:
\begin{equation}
\mathbf{K}_e = \int_{V_e} \mathbf{B}_e^\mathsf{T} \mathbf{D}_e \mathbf{B}_e \, dV,
\label{eq:element_stiffness}
\end{equation}
where \( V_e \) is the volume of the element, \( \mathbf{B}_e \) is the strain-displacement matrix derived from the element shape functions, and \( \mathbf{D}_e \) is the material stiffness matrix, which encodes the linear elastic constitutive law. For isotropic materials, \( \mathbf{D}_e \in \mathbb{R}^{6 \times 6} \) is defined as:
\begin{equation}
\mathbf{D}_e =
\begin{bmatrix}
\lambda + 2\mu & \lambda & \lambda & 0 & 0 & 0 \\
\lambda & \lambda + 2\mu & \lambda & 0 & 0 & 0 \\
\lambda & \lambda & \lambda + 2\mu & 0 & 0 & 0 \\
0 & 0 & 0 & \mu & 0 & 0 \\
0 & 0 & 0 & 0 & \mu & 0 \\
0 & 0 & 0 & 0 & 0 & \mu
\end{bmatrix},
\label{eq:material_matrix}
\end{equation}
where \( \lambda \) and \( \mu \) are Lamé parameters, computed from \( E \) and \( \nu \) via \( \lambda = \frac{E \nu}{(1+\nu)(1-2\nu)} \), \( \mu = \frac{E}{2(1+\nu)} \).

The global stiffness matrix is then obtained by summing the local contributions over all elements through standard finite element assembly:
\begin{equation}
\mathbf{K} = \sum_e \mathbf{A}_e^\mathsf{T} \mathbf{K}_e \mathbf{A}_e,
\label{eq:global_stiffness}
\end{equation}
where \( \mathbf{A}_e \) is the element-to-global mapping operator.

The quadratic form \( \mathbf{u}_{\Omega}^\mathsf{T}  \mathbf{K} \mathbf{u}_{\Omega} \) quantifies the total strain energy within the deformable domain. The stiffness matrix \( \mathbf{K} \) encodes the tissue’s mechanical properties and governs the propagation of stress and strain, promoting physically coherent deformations throughout the organ.

\subsection{Forward Problem}
In FE modeling, the forward problem refers to computing the displacements of free (unconstrained) nodes within the volumetric domain \( \Omega \), given prescribed displacements at constrained boundary nodes. When the constraints span the entire model boundary \( \partial \Omega \), this process represents mapping known boundary displacements to interior deformation, corresponding to the reverse direction illustrated in Fig.~\ref{fig:interpolation_matrix}.

Let the global displacement vector of the volumetric mesh \( \Omega \) be denoted as \( \mathbf{u}_{\Omega} \in \mathbb{R}^{3n_v} \), which is partitioned into:
\begin{equation}
\mathbf{u}_{\Omega} =
\begin{bmatrix}
\mathbf{u}_f \\
\mathbf{u}_c
\end{bmatrix},
\label{eq:partition}
\end{equation}
where \( \mathbf{u}_f \in \mathbb{R}^{3n_f} \) denotes the displacements of the free nodes, and \( \mathbf{u}_c \in \mathbb{R}^{3n_c} \) denotes the displacements of the constrained nodes, with \( n_f + n_c = n_v \).

Substituting this partition into the static equilibrium equation in Eq.~\eqref{eq:static_equilibrium} yields the following block matrix system:
\begin{equation}
\begin{bmatrix}
\mathbf{K}_{ff} & \mathbf{K}_{fc} \\
\mathbf{K}_{cf} & \mathbf{K}_{cc}
\end{bmatrix}
\begin{bmatrix}
\mathbf{u}_f \\
\mathbf{u}_c
\end{bmatrix}
=
\begin{bmatrix}
\mathbf{f}_f \\
\mathbf{f}_c
\end{bmatrix},
\label{eq:block_system}
\end{equation}
where the stiffness submatrices \(\mathbf{K}_{ff}\), \(\mathbf{K}_{fc}\), etc., represent the interactions among free and constrained nodes, and \(\mathbf{f}_f\), \(\mathbf{f}_c\) are the corresponding nodal force vectors.

Assuming that no external force is applied to the free nodes (i.e., \(\mathbf{f}_f = \mathbf{0}\)), the displacement of free nodes is computed by solving the reduced system:
\begin{equation}
\mathbf{u}_f = -\mathbf{K}_{ff}^{-1} \mathbf{K}_{fc} \mathbf{u}_c.
\label{eq:forward_solution}
\end{equation}
The complete displacement vector \( \mathbf{u}_{\Omega} \) is then reconstructed by concatenating the free and constrained components as described in Eq.~\eqref{eq:partition}.

\subsection{Correspondence Matrix}
The correspondence matrix \( \mathbf{C} \in \mathbb{R}^{n \times m} \) establishes the mapping between points in the preoperative organ model \( \mathcal{X} = \{\mathbf{x}_i\}_{i=1}^n \) and the intraoperative point cloud \( \mathcal{Y} = \{\mathbf{y}_j\}_{j=1}^m \). It encodes the geometric relationships between the two surfaces, facilitating their alignment and enabling the estimation of the deformation field on observable regions. Depending on the formulation, \( \mathbf{C} \) can represent either probabilistic or deterministic correspondences.

In its probabilistic form, \( \mathbf{C} \) is a soft correspondence matrix where each entry \( \mathbf{C}_{i,j} \in [0,1] \) indicates the likelihood that \( \mathbf{x}_i \in \mathcal{X} \) corresponds to \( \mathbf{y}_j \in \mathcal{Y} \). In contrast, a binary matrix \( \mathbf{C} \in \{0,1\}^{n \times m} \) represents deterministic one-to-one matching, where \( \mathbf{C}_{i,j} = 1 \) denotes a strict correspondence between \( \mathbf{x}_i \) and \( \mathbf{y}_j \). In this study, we adopt the binary formulation to achieve precise correspondence.

To construct the binary matrix \( \mathbf{C} \), a soft correspondence matrix is first predicted using either learning-based or geometry-based approaches \cite{han2024review}. The soft matrix is then converted into a binary form by solving a linear assignment problem (LAP), enforcing strict one-to-one matching:
\begin{equation}
\sum_{j} \mathbf{C}_{i,j} = 1, \forall i; \ \sum_{i} \mathbf{C}_{i,j} = 1, \forall j; \ \text{and\ } \mathbf{C}_{i,j} \in \{0, 1\}.
\end{equation}
However, in the presence of partial overlaps between \( \mathcal{X} \) and \( \mathcal{Y} \), such strict matching may yield spurious correspondences due to missing regions or noise. To mitigate this, additional outlier pruning is performed after LAP solving to remove low-confidence matches and refine the final matrix \( \mathbf{C} \). Notably, this approach does not rely on any pre-existing rigid registration between the two point sets.

Alternatively, when a reliable rigid alignment \( \mathcal{R} \) has already been obtained, correspondences can be directly inferred via mutual nearest neighbors with geometric filtering:
\begin{equation}
\mathbf{C}_{i,j} =
\begin{cases}
1, & \text{NN}(\mathcal{R}(\mathbf{x}_i), \mathcal{Y}) = \mathbf{y}_j, \text{NN}(\mathcal{X}, \mathcal{R}^{-1} (\mathbf{y}_j)) = \mathbf{x}_i, \,  \\
   & \quad \quad \quad \text{and } \|\mathcal{R}(\mathbf{x}_i) - \mathbf{y}_j\|_2 < r, \\
0, & \text{otherwise},
\end{cases}
\end{equation}
where \( \mathcal{R}(\cdot) \) denotes the rigid transformation, \( \text{NN}(\cdot) \) is the nearest neighbor operator, and \( r \) is a distance threshold to filter out unreliable matches. This strategy provides an efficient correspondence estimation under pre-aligned conditions.

In our implementation, the correspondence matrix for the synthetic liver dataset is initially established using UTOPIC~\cite{chen2022utopic} and refined with GraphSCNet~\cite{qin2023deep} to filter outliers. For other datasets, correspondences are established using mutual nearest neighbors.

Once reliable correspondences are established, the deformation characteristics over the observable region can be quantified by computing point-wise residuals between matched pairs:
\begin{equation}
\mathbf{r}_i = \sum_{j=1}^{m} \mathbf{C}_{i,j} \left( \mathbf{y}_j - \mathbf{x}_i \right),
\label{eq:residual}
\end{equation}
where \( \mathbf{r}_i \in \mathbb{R}^3 \) captures the local discrepancy between the preoperative point \( \mathbf{x}_i \) and its matched intraoperative observations. This residual serves as an indicator of local surface deformation, reflecting deviations between the preoperative model and intraoperative observations. In this sense, the correspondence matrix \( \mathbf{C} \) enables a spatially resolved representation of deformation over the observable surface, which can be further leveraged to inform volumetric biomechanical modeling.

\begin{table}[htbp]
\centering
\caption{Mathematical Notations}
\begingroup
\renewcommand{\arraystretch}{1.2}
\begin{tabularx}{0.48\textwidth}{p{0.8cm}Xp{1.4cm}}
\toprule
\textbf{Symbol} & \textbf{Description} & \textbf{Dimension} \\
\midrule
\multicolumn{3}{l}{\textbf{Geometric Symbols}} \\
\( \mathcal{X}\) & Preoperative organ surface \(\{ \mathbf{x}_i \}_{i=1}^n\) & \( \mathbb{R}^{n\times3} \) \\
\( \mathcal{Y}\) & Intraoperative surface point cloud \(\{ \mathbf{y}_i \}_{i=1}^m\) & \(\mathbb{R}^{m\times3}\) \\
\( \mathcal{X}' \) & Deformed organ surface \(\{ \mathbf{x}_i+ \mathbf{u}_i\}_{i=1}^n\) & \( \mathbb{R}^{n\times3} \) \\
\( \Omega \) & Volumetric tetrahedral mesh & — \\
\( \partial \Omega \) & Surface boundary of \( \Omega \) & — \\
\( \mathcal{X}_\Omega \) & Volumetric mesh nodes \(\{\mathbf{x}_{\Omega,i}\}_{i=1}^{n_v} \) & \(\mathbb{R}^{n_v\times3}\) \\ 
\midrule
\multicolumn{3}{l}{\textbf{Displacement Field Symbols}} \\
\( \mathbf{u}_i \) & Displacement of surface point \( \mathbf{x}_i \) & \( \mathbb{R}^3 \) \\
\( \mathbf{u} \) & Global boundary displacement field & \( \mathbb{R}^{3n} \) \\
\( \mathbf{u}_\Omega \) & Volumetric displacement field & \( \mathbb{R}^{3n_v} \) \\
\( \mathbf{u}_\Omega' \) & Incremental displacement w.r.t. \(\mathcal{X}'\) & \( \mathbb{R}^{3n_v} \) \\ 
\midrule
\multicolumn{3}{l}{\textbf{Network and Encoding Symbols}} \\
\( \Theta_l \) & MLP at pyramid level \( l \) & — \\
\( \Gamma_l(\cdot) \) & Sinusoidal positional encoder & \( \mathbb{R}^6 \) \\
\( \gamma_i^l \) & Encoded coordinate at level \( l \) & \( \mathbb{R}^6 \) \\
\( c^l(\gamma) \) & Confidence score (range: \( (0,1) \)) & $\mathrm{scalar}$ \\
\( \mathbf{a}^l(\gamma) \) & Directional scaling vector & \( \mathbb{R}^3 \) \\ 
\midrule
\multicolumn{3}{l}{\textbf{Matrices and Operators}} \\
\( \mathbf{K} \) & Stiffness matrix & \( \mathbb{R}^{3n_v \times 3n_v} \) \\
\( \Phi \) & Interpolation matrix & \( \mathbb{R}^{3n \times 3n_v} \) \\
\( \mathbf{C} \) & Binary correspondence matrix & \( \{0,1\}^{n \times m} \) \\
\( \tilde{\mathbf{C}} \) & Kronecker-expanded correspondence & \( \{0,1\}^{3n \times 3m} \) \\
\( \mathbf{P} \) & Diagonal matrix of correspondence weights & \( \mathbb{R}^{3n \times 3n} \) \\ 
\midrule
\multicolumn{3}{l}{\textbf{Miscellaneous Parameters}} \\
\( \sigma^2 \) & Residual-based regularization weight & $\mathrm{scalar}$ \\
\( \beta \) & Tikhonov weight of biomechanical energy & $\mathrm{scalar}$ \\
\( \lambda, \mu \) & Lamé material constants & $\mathrm{scalars}$ \\
\bottomrule
\end{tabularx}
\endgroup
\label{tab:notation}
\end{table}

\section{Methods}
\subsection{Overview}
\begin{figure*}[!t]
\centering
\includegraphics[width=0.96\textwidth]{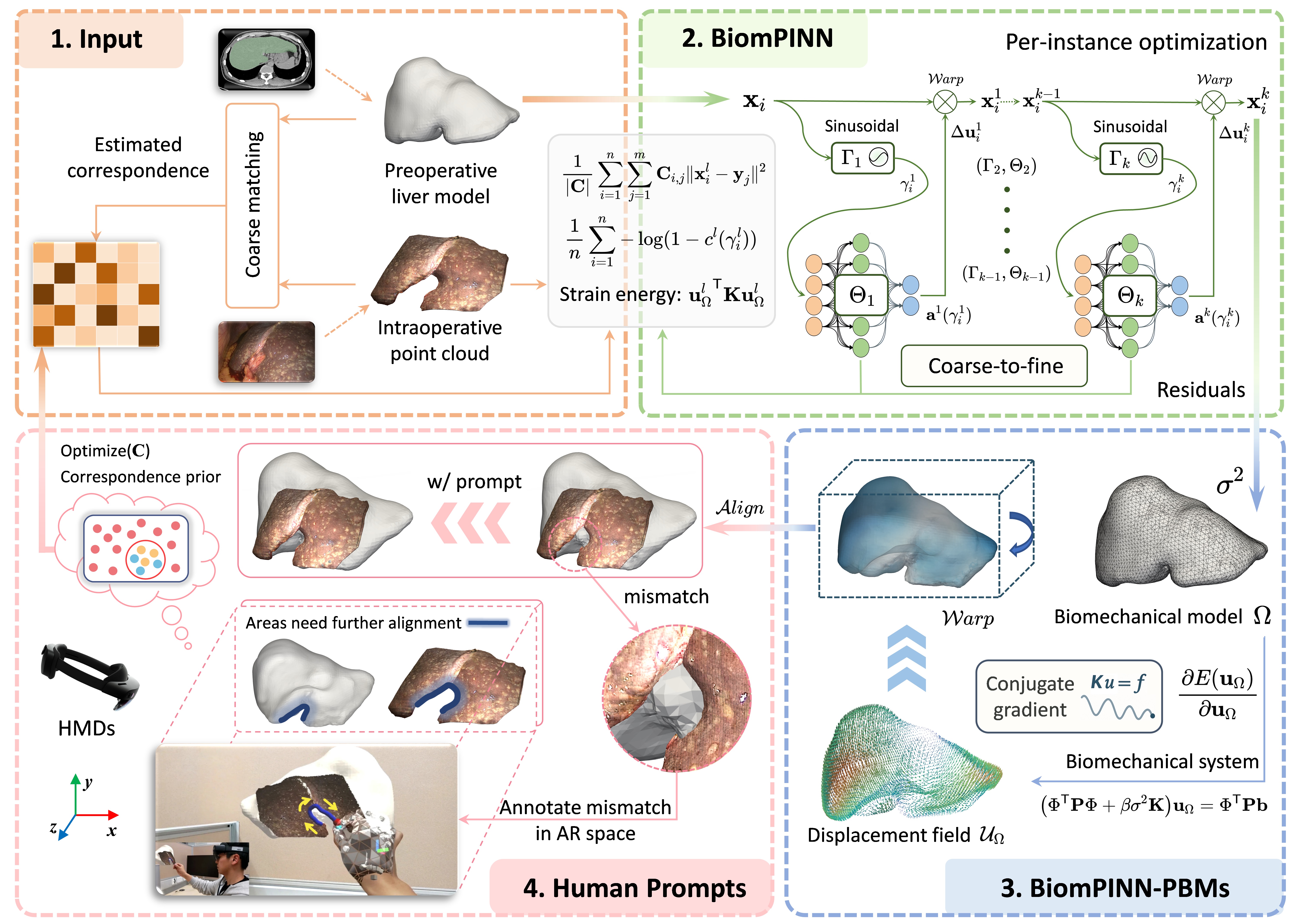}
\caption{Overview of the proposed interactive deformation modeling framework for AR-guided surgical navigation. The process consists of four sequential steps: (1) input of the preoperative model and intraoperative point cloud with initial coarse matching; (2) deformation estimation using BiomPINN, a physics-integrated, per-instance optimized network; (3) refinement through integration with a biomechanical finite element model (BiomPINN-PBMs); and (4) interactive human-in-the-loop correction using line-drawn prompts via AR interfaces. }
\label{fig:framework}
\vspace{-10pt}
\end{figure*}

Fig. \ref{fig:framework} illustrates the overall pipeline of the proposed interactive organ deformation modeling framework. The process begins with three primary inputs: a preoperative liver model reconstructed from CT scans, an intraoperative point cloud derived from laparoscopic images, and their correspondences. These inputs drive the BiomPINN algorithm, which deforms the preoperative model based on the correspondences while integrating an FE model as a constraint to ensure physically plausible deformations. Building upon the output of BiomPINN, a biomechanical system is formulated and efficiently solved in a single-pass optimization to further improve deformation accuracy.

To address instances of unsatisfactory results, the framework integrates an AR-based interactive system that allows surgeons to actively participate in the deformation process. Surgeons can provide prompts in AR space to specify regions of the organ requiring further alignment. The framework treats these prompts as correspondence priors to correct mismatched regions, leveraging biomechanics to propagate local corrections throughout the entire organ model.

The key mathematical notations used throughout this section are summarized in Table~\ref{tab:notation}.

\subsection{BiomPINN}
To estimate non-rigid deformation between the preoperative organ model and intraoperative observations, we propose a framework that incorporates a Multi-Layer Perceptron (MLP)-based hierarchical deformation pyramid~\cite{li2022non} with a biomechanical regularizer derived from an FE model, termed \textit{BiomPINN}. Unlike conventional learning-based approaches that rely on an offline training–testing paradigm, BiomPINN adopts a \textit{per-instance optimization} strategy: for each deformation modeling task, network parameters are initialized and optimized from scratch. This design enables the model to adapt to the deformation observed in each individual case, without relying on generalization across datasets.

BiomPINN models the displacement field \( \mathbf{u} \in \mathbb{R}^{3n} \) over the organ boundary \( \partial \Omega \) using a hierarchical pyramid structure with \( k \) levels. The pyramid is defined as \( \Delta = \{ (\Gamma_l, \Theta_l) \mid l = 1, \dots, k \} \), where each level \( l \) includes a sinusoidal positional encoding function \( \Gamma_l \) and an optimizable MLP \( \Theta_l \). Optimization proceeds sequentially across levels: for each level \( l \), only the parameters of \( \Theta_l \) are updated, while all previous networks \( \Theta_1, \dots, \Theta_{l-1} \) are frozen.

Each MLP \( \Theta_l \) predicts an incremental displacement \( \Delta \mathbf{u}_i^l \in \mathbb{R}^3 \) for each boundary point \( \mathbf{x}_i \in \mathcal{X}\), based on the deformed coordinates from the previous level. The cumulative displacement and updated position at level \( l \) are given by:
\begin{equation}
\mathbf{u}_i^l = \Delta \mathbf{u}_i^1 + \cdots + \Delta \mathbf{u}_i^l,
\quad
\mathbf{x}_i^l = \mathbf{x}_i + \mathbf{u}_i^l.
\label{eq:updated coordinates}
\end{equation}
The updated surface at level \( l \) is denoted as \( \mathcal{X}^l = \{ \mathbf{x}_i^l \}_{i=1}^n \). After the final pyramid level \( k \), the predicted deformed surface is denoted as \( \mathcal{X}' = \mathcal{X} + \mathbf{u} \), where \( \mathbf{u} = \sum_{l=1}^k \Delta \mathbf{u}^l \) represents the total boundary displacement.

To encode geometry, each input coordinate is passed through a level-specific sinusoidal embedding:
\begin{equation}
\Gamma_l(\mathbf{x}) = \left[\sin(4^l \pi \mathbf{x}), \cos(4^l \pi \mathbf{x})\right] \in \mathbb{R}^6,
\end{equation}
with the encoded feature denoted as \( \gamma_i^l = \Gamma_l(\mathbf{x}_i^{l-1}) \). For the first level (\( l = 1 \)), we define \( \mathbf{x}_i^0 = \mathbf{x}_i \) as the original surface point. The network \( \Theta_l \) then outputs a confidence score \( c^l(\gamma_i^l) \in (0,1) \) and a directional scaling vector \( \mathbf{a}^l(\gamma_i^l) \in \mathbb{R}^3 \), which together define the incremental displacement:
\begin{equation}
\Delta \mathbf{u}_i^l = c^l(\gamma_i^l) \cdot \mathbf{a}^l(\gamma_i^l) \circ \mathbf{x}_i^{l-1},
\label{eq:incremental displacement}
\end{equation}
where \( \circ \) denotes element-wise multiplication.

To evaluate biomechanical plausibility, the predicted boundary displacement at level \( l \), denoted by \( \mathbf{u}^l = [\mathbf{u}_1^l, \dots, \mathbf{u}_n^l]^\mathsf{T} \), is projected to the volumetric mesh \( \Omega \) using the interpolation matrix \( \Phi^\mathsf{T} \in \mathbb{R}^{3n_v \times 3n} \):
\begin{equation}
\mathbf{u}_\Omega^l = \Phi^\mathsf{T} \mathbf{u}^l.
\label{eq:interpolation_reverse}
\end{equation}
Here, boundary displacements are preserved while internal node displacements are set to zero.

\textbf{Cost function.} At each pyramid level \( l \), the total loss comprises three components:
\begin{equation}
\mathcal{L}^l = \mathcal{L}_{\text{align}}^l + \lambda_1 \mathcal{L}_{\text{rigid}}^l + \lambda_2 \mathcal{L}_{\text{fem}}^l.
\label{eq:loss}
\end{equation}

\noindent \textit{Geometric alignment term} (\( \mathcal{L}_{\text{align}}^l \)):  
This term enforces alignment between the deformed surface \( \mathcal{X}^l = \{ \mathbf{x}_i^l \}_{i=1}^n \) and the intraoperative point cloud \( \mathcal{Y} = \{ \mathbf{y}_j \}_{j=1}^m \), based on a binary correspondence matrix \( \mathbf{C} \in \{0,1\}^{n \times m} \):
\begin{equation}
\mathcal{L}_{\text{align}}^l = \frac{1}{|\mathbf{C}|} \sum_{i=1}^{n} \sum_{j=1}^{m} \mathbf{C}_{i,j} \| \mathbf{x}_i^l - \mathbf{y}_j \|^2.
\end{equation}

\noindent \textit{Geometric regularization term} (\( \mathcal{L}_{\text{rigid}}^l \)):  
This term discourages large displacements in regions with low confidence scores, thereby promoting locally rigid behavior:
\begin{equation}
\mathcal{L}_{\text{rigid}}^l = \frac{1}{n} \sum_{i=1}^{n} -\log(1 - c^l(\gamma_i^l)).
\end{equation}

\noindent \textit{Biomechanical regularization term} (\( \mathcal{L}_{\text{fem}}^l \)):  
To penalize non-physical deformation, this term computes the FE model-based strain energy over the projected volumetric displacement field:
\begin{equation}
\mathcal{L}_{\text{fem}}^l = \alpha_l \, {\mathbf{u}_\Omega^l}^\mathsf{T} \mathbf{K} \mathbf{u}_\Omega^l, \quad \alpha_l = \frac{k - l}{2k}.
\end{equation}
Since the displacements of internal nodes are set to zero during interpolation Eq.~\eqref{eq:interpolation_reverse}, this regularizer constrains only the boundary nodes, thereby acting as a system of virtual springs between neighboring surface points. Through the element-wise coupling encoded in the stiffness matrix \(\mathbf{K}\), it suppresses locally inconsistent motion while enabling the propagation of deformation from observed to unobserved regions, thereby promoting globally coherent behavior.

\textbf{Per-instance optimization paradigm.}  
Each deformation modeling task is optimized independently, with no parameter sharing across cases. The optimization proceeds sequentially across the pyramid levels: at each level \( l \), the parameters of network \( \Theta_l \) are optimized, while those of all earlier networks \( \Theta_1, \dots, \Theta_{l-1} \) remain frozen. This coarse-to-fine refinement strategy enables the model to capture large-scale global deformation at early levels and progressively refine local structure in later stages. The overall optimization process is summarized in Algorithm~\ref{alg:biompinn}, which details the hierarchical parameter updates and deformation refinement steps.

\begin{algorithm}[htbp]
\caption{Per-instance Optimization for BiomPINN}
\label{alg:biompinn}
\begin{algorithmic}[1]
\State \textbf{Input:} Preoperative surface $\mathcal{X} \subset \partial \Omega$, intraoperative point cloud $\mathcal{Y}$, correspondence matrix $\mathbf{C}$, stiffness matrix $\mathbf{K}$
\State \textbf{Output:} Predicted displacement $\mathbf{u}$, deformed surface $\mathcal{X}'$
\State Initialize cumulative displacement $\mathbf{u}_i^0 \gets \mathbf{0},\ \forall \mathbf{x}_i \in \mathcal{X}$
\State Initialize current surface $\mathcal{X}^0 \gets \mathcal{X}$

\For{$l = 1$ to $k$} \Comment{Iterate over pyramid levels}
    \State Initialize network parameters $\Theta_l$ (e.g., Xavier uniform)
    \State Freeze all previous networks $\Theta_1, \dots, \Theta_{l-1}$
    
    \ForAll{points $\mathbf{x}_i^{l-1} \in \mathcal{X}^{l-1}$}
        \State Compute encoded coordinate $\gamma_i^l \gets \Gamma_l(\mathbf{x}_i^{l-1})$
        \State Predict deformation parameters $\Theta_l(\gamma_i^l)$
        \State Compute displacement vector $\Delta \mathbf{u}_i^l$ \Comment{Eq.~\eqref{eq:incremental displacement}}
        \State Update cumulative displacement $\mathbf{u}_i^l$ \Comment{Eq.~\eqref{eq:updated coordinates}}
        \State Update position $\mathbf{x}_i^l \gets \mathbf{x}_i + \mathbf{u}_i^l$
    \EndFor

    \State Assemble surface displacement: $\mathbf{u}^l \gets [\mathbf{u}_1^l, \dots, \mathbf{u}_n^l]^\mathsf{T}$
    \State Project to volume: $\mathbf{u}_\Omega^l \gets \Phi^\mathsf{T} \mathbf{u}^l$ \Comment{Eq.~\eqref{eq:interpolation_reverse}}
    \State Update deformed surface: $\mathcal{X}^l \gets \mathcal{X} + \Phi \mathbf{u}_\Omega^l$
    \State Compute loss $\mathcal{L}^l$ \Comment{Eq.~\eqref{eq:loss}}
    \State Update network $\Theta_l$ via gradient descent on $\mathcal{L}^l$
\EndFor

\State Compute total displacement: $\mathbf{u} \gets \sum_{l=1}^k \mathbf{u}^l$
\State Update surface: $\mathcal{X}' \gets \mathcal{X} + \mathbf{u}$
\State \Return $\mathbf{u}$,  $\mathcal{X}'$
\end{algorithmic}
\end{algorithm}

By integrating geometry-aware learning with physics-based constraints, BiomPINN produces physically plausible deformation fields from sparse surface observations, making it well-suited for real-world surgical scenarios where volumetric supervision is typically unavailable.

\subsection{BiomPINN-PBMs Integration}
To leverage both the predictive efficiency of learning-based models and the physical plausibility of biomechanical simulation, we propose an integrated framework, termed \textit{BiomPINN-PBMs Integration}. In this formulation, the surface deformation field \( \mathcal{U} \), predicted by BiomPINN, is used to directly determine the regularization weight in a PBM, thereby avoiding costly iterative estimation and enabling a one-pass optimization to the volumetric deformation.

The PBM estimates the volumetric deformation field \( \mathbf{u}_\Omega \in \mathbb{R}^{3n_v} \) by minimizing a composite energy function comprising elastic and data fidelity terms:
\begin{equation}
\begin{split}
E(\mathbf{u}_{\Omega}) &= \frac{\beta}{2} \mathbf{u}_{\Omega}^\mathsf{T} \mathbf{K} \mathbf{u}_{\Omega} \\
&\quad + \frac{1}{2\sigma^2}\sum_{i=1}^{3n} \sum_{j=1}^{3m} \tilde{\mathbf{C}}_{i,j} \|\Phi_i(\mathbf{x}_{\Omega} + \mathbf{u}_{\Omega}) - \Vec{\mathbf{y}}_j\|^2,
\end{split}
\label{eq:pbm_energy}
\end{equation}
where the first term quantifies the elastic strain energy stored in the deformed volumetric mesh \(\Omega\), scaled by a Tikhonov weight \(\beta\). The second term measures the deviation between the deformed source mesh \(\partial\Omega\) and the intraoperative target point cloud \(\mathcal{Y}\). Here, \(\tilde{\mathbf{C}}=\text{kron}(\mathbf{C},I_{3\times3})\in\{0, 1\}^{3n \times 3m}\) is is the Kronecker product of a binary correspondence matrix 
\(\mathbf{C}\) and the identity matrix \(I_{3\times3}\), allowing pointwise matching in vectorized 3D space. Each \(\Phi_i \in \mathbb{R}^{1\times3n_v}\) is the \(i\)-th row-block of the interpolation matrix \(\Phi \in \mathbb{R}^{3n\times3n_v}\), projecting volumetric displacements to the surface, and \( \Vec{\mathbf{y}} \in \mathbb{R}^{3m} \) is the vectorized intraoperative point cloud.

In this setting, the parameter \( \sigma^2 \) controls the trade-off between physical plausibility and geometric alignment. Conventionally, determining an appropriate \( \sigma^2 \) requires iterative optimization, such as expectation-maximization, where the energy function is minimized repeatedly across different values of \( \sigma^2 \). To avoid this costly process, we adopt a residual-based formulation for computing \( \sigma^2 \) directly using the prediction from BiomPINN:
\begin{equation}
\sigma^2 = \frac{1}{|\tilde{\mathbf{C}}|} 
\sum_{i=1}^{3n} \sum_{j=1}^{3m} \tilde{\mathbf{C}}_{i,j} \left\| \Phi_i (\mathbf{x}_\Omega + \mathbf{u}_\Omega) - \Vec{\mathbf{y}}_j \right\|^2.
\label{eq:sigma_estimate}
\end{equation}
At this stage, the deformation field \(\mathbf{u}_\Omega\) is computed by projecting the boundary displacement \(\mathbf{u}\), predicted by BiomPINN, to the volumetric mesh using the interpolation operator defined in Eq.~\eqref{eq:interpolation_reverse}.

Once \(\sigma^2\) is computed, it is substituted back into Eq.~\eqref{eq:pbm_energy} to solve for the refined volumetric deformation \(\mathcal{U}_\Omega\). Taking the derivative of the derivative of \(E(\mathbf{u}_{\Omega})\) with respect to \(\mathbf{u}_{\Omega}\) and setting it to zero leads to the following normal equations:
\begin{equation}
\frac{\partial E(\mathbf{u}_{\Omega})}{\partial \mathbf{u}_{\Omega}} = \beta\mathbf{K}\mathbf{u}_{\Omega} + \frac{1}{\sigma^2}\Phi^\mathsf{T}\mathbf{P}[\Phi(\mathbf{x}_{\Omega}+\mathbf{u}_{\Omega})-\Tilde{\mathbf{C}}\Vec{\mathbf{y}}] = 0,
\end{equation}
which yields the linear system:
\begin{equation}
\left( \Phi^\mathsf{T} \mathbf{P} \Phi + \beta \sigma^2 \mathbf{K} \right) \mathbf{u}_\Omega 
= \Phi^\mathsf{T} \mathbf{P} \mathbf{b},
\label{eq:pbm_linear_system}
\end{equation}
where \( \mathbf{P} = \mathrm{diag}(\tilde{\mathbf{C}} \mathbf{1}) \in \mathbb{R}^{3n \times 3n} \) is a diagonal matrix that accumulates correspondence weights from \( \tilde{\mathbf{C}} \), and \( \mathbf{b} = -\Phi \mathbf{x}_\Omega + \tilde{\mathbf{C}} \Vec{\mathbf{y}} \) denotes the residual offset between the predicted and observed surface displacements. This system can be solved via conjugate gradient methods, yielding the final deformation field \( \mathcal{U}_\Omega \) over the volumetric mesh \( \Omega \).

Through this integration, the deformation field predicted by BiomPINN not only serves as a surface alignment prior but also guides the estimation of the optimal regularization strength \( \sigma^2 \) within the PBM. This removes the need for iterative estimation and allows for a single-pass solution of Eq.~\eqref{eq:pbm_linear_system}. As a result, the BiomPINN-PBMs integration achieves accurate and physically plausible volumetric deformation while maintaining computational efficiency suitable for intraoperative deployment.

\subsection{Interactive Deformation Modeling Framework}

To further enhance the accuracy and reliability of deformation modeling, we propose an \textit{Interactive Deformation Modeling Framework} that incorporates a human-in-the-loop mechanism. This framework empowers surgeons to directly intervene when the predicted deformation fails to sufficiently align the preoperative model \( \mathcal{X} \) with the intraoperative observations \( \mathcal{Y} \). Through intuitive AR-based annotations, the system allows targeted refinement of the deformation field based on expert feedback.

\textbf{Core mechanism.}  
The key insight behind this framework is that many local misalignments originate from incorrect or missing point correspondences between the preoperative and intraoperative surfaces. Rather than directly modifying the deformation field, we focus on correcting the underlying correspondence matrix \( \mathbf{C} \), which governs surface alignment. This is achieved through an interactive prompt design, wherein surgeons specify target regions via line-based annotations—\( \mathcal{L}_{\mathcal{X}} \) on the deformed preoperative surface and \( \mathcal{L}_{\mathcal{Y}} \) on the intraoperative point cloud—indicating areas that should correspond but are currently misaligned.
To operationalize these prompts, the drawn lines are first sampled into discrete 3D points, which are then expanded into local point sets \( \mathcal{X}_m \subseteq \mathcal{X}' \) and \( \mathcal{Y}_m \subseteq \mathcal{Y} \) by retrieving their five nearest neighbors from the respective meshes. These sets form spatially coherent regions to guide refinement. To reduce spatial mismatch, an iterative closest point (ICP) algorithm is applied to estimate a rigid transformation \( \mathcal{R} \in \mathrm{SE}(3) \) that aligns \( \mathcal{X}_m \) to \( \mathcal{Y}_m \). The correspondence matrix is then updated through mutual nearest neighbor matching:
\begin{equation}
\mathbf{C}_{i,j} =
\begin{cases}
1, & \text{if } \mathrm{NN}(\mathcal{R}(\mathbf{x}_i), \mathcal{Y}_m) = \mathbf{y}_j, \quad \mathbf{x}_i \in \mathcal{X}_m, \\
0, & \text{otherwise}.
\end{cases}
\label{eq:interactive_corr}
\end{equation}

\begin{figure}[!t]
\centerline{\includegraphics[width=\columnwidth]{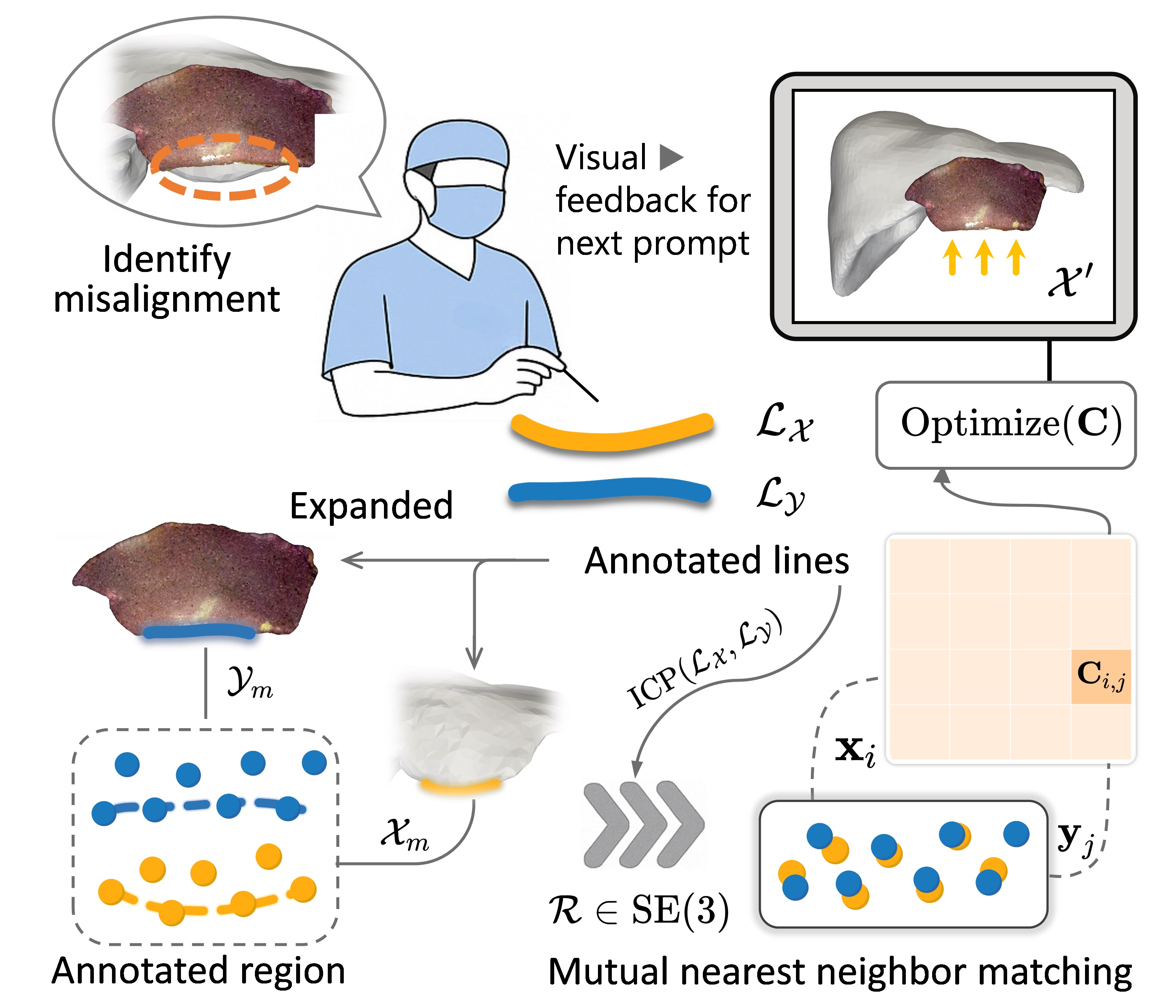}}
\caption{Interactive deformation modeling workflow. User-annotated misaligned regions are expanded into local surface patches, aligned via rigid ICP, and used to update the correspondence matrix via mutual nearest neighbor matching. The deformation field is then re-optimized and rendered, providing intuitive visual feedback for assessing alignment quality and guiding further user input if necessary.}
\label{fig:human_prompt}
\vspace{-10pt}
\end{figure}

\textbf{Prompt-guided deformation update.}  
With the updated correspondence matrix, the deformation field is re-optimized using the deformation modeling framework to produce an updated volumetric displacement field \( \mathbf{u}_\Omega' \).  
Note that \( \mathbf{u}_\Omega' \) is defined with respect to the current deformed surface \( \mathcal{X}' \), rather than the original model \( \mathcal{X} \). This incremental formulation ensures that each refinement is applied locally on top of the previously deformed configuration. The surface is then updated using the interpolation matrix \( \Phi \):
\begin{equation}
\mathcal{X}' \leftarrow \mathcal{X}' + \Phi \mathbf{u}_\Omega', \quad \mathbf{u}_\Omega \leftarrow \mathbf{u}_\Omega+\mathbf{u}_\Omega'.
\label{eq:update}
\end{equation}
The updated surface \( \mathcal{X}' \) is rendered for inspection, enabling intuitive evaluation of surface alignment quality and guiding additional rounds of user input if necessary.

\begin{algorithm}[!t]
\caption{Interactive Deformation Modeling}
\label{alg:interactive}
\begin{algorithmic}[1]
\State \textbf{Input:} Deformation field \( \mathbf{u}_\Omega \), undeformed surface \( \mathcal{X} \), point cloud \( \mathcal{Y} \), correspondence matrix \( \mathbf{C} \)
\State \textbf{Output:} Updated deformation \(\mathbf{u}_\Omega\), deformed surface \( \mathcal{X}' \)
\State Compute deformed surface: \( \mathcal{X}' \leftarrow \mathcal{X} + \Phi \mathbf{u}_\Omega \)
\State Render \( \mathcal{X}' \) for surgeon inspection
\While{surgeon identifies misalignment}
    \State Surgeon draws annotations: \( \mathcal{L}_\mathcal{X}, \mathcal{L}_\mathcal{Y} \)
    \State Sample lines and expand to local regions \( \mathcal{X}_m, \mathcal{Y}_m \)
    \State Estimate rigid transformation \( \mathcal{R} \leftarrow \mathrm{ICP}(\mathcal{L}_\mathcal{X}, \mathcal{L}_\mathcal{Y}) \)
    \State Apply transformation: \( \mathcal{X}_m \leftarrow \mathcal{R}(\mathcal{X}_m) \)
    \State Update correspondence matrix \( \mathbf{C} \)\Comment{Eq.~\eqref{eq:interactive_corr}}
    \State Optimize deformation: \( \mathbf{u}_\Omega' \leftarrow \mathrm{Optimize}(\mathbf{C}) \)
    \State Update deformation field and surface \Comment{Eq.~\eqref{eq:update}}
    \State Render updated \( \mathcal{X}' \) for feedback
\EndWhile
\State \Return \( \mathbf{u}_\Omega, \mathcal{X}' \)
\end{algorithmic}
\end{algorithm}

This interactive framework enables expert feedback to be seamlessly integrated into the deformation pipeline, particularly in anatomically ambiguous regions or when visual cues are insufficient. By refining correspondences rather than directly editing displacements, the approach remains compatible with existing optimization routines while preserving biomechanical consistency. The biomechanical regularizer embedded in the deformation modeling framework further ensures that refinements are not restricted to annotated regions alone but propagate coherently across the volumetric mesh \( \Omega \), resulting in anatomically consistent and physically plausible deformation across the organ. A step-by-step description of this human-in-the-loop refinement procedure is provided in Algorithm~\ref{alg:interactive}.

\section{EXPERIMENTAL SET-UP}
To evaluate the performance of the proposed algorithms and framework, three experiments were conducted: 1) biomechanical simulations to assess the algorithm's approximation of numerical results from the FE-based approach, 2) quantitative evaluation of volumetric accuracy using a 3D-printed silicone liver phantom, and 3) qualitative and quantitative evaluation on human in-vivo laparoscopic cases.
\subsection{Datasets}
\subsubsection{Synthetic dataset}
To evaluate how well our method approximates the numerical results of an FE-based biomechanical simulation, a synthetic dataset was constructed for three organs: liver, kidney, and prostate. The dataset generation followed the methodology introduced in~\cite{pfeiffer2020non}.

Organ-specific simulation parameters are summarized in Table~\ref{tab:organ-simulation-parameters}. Starting from preoperative surface meshes, volumetric FE models were constructed using CT-derived geometries. A neo-Hookean hyperelastic material model was assigned to represent soft tissue properties for simulating organ deformations. For all organs, Young's modulus was randomly sampled between 2~kPa and 5~kPa, and the Poisson's ratio was fixed at 0.35. Depending on the organ, one to three forces were applied to random surface regions, with organ-specific upper bounds on force magnitude. Zero-displacement boundary conditions (BC) were imposed on selected surface regions, with radii drawn from organ-dependent ranges. These simulations were implemented using the open-source FE solver Elmer~\cite{malinen2013elmer}.

To emulate intraoperative observations, partial surface point clouds were extracted from the deformed meshes by randomly sampling visible patches. Additionally, vertex-wise perturbations were introduced to simulate intraoperative noises, with displacement magnitudes capped by organ-specific thresholds. A total of 300, 360, and 160 deformation instances were generated for the liver, kidney, and prostate, respectively, enabling evaluation of surface-to-surface registration accuracy across diverse anatomical and biomechanical settings.

\begin{table}[!t]
\centering
\caption{Organ-specific simulation parameters used for generating synthetic deformation datasets.}
\resizebox{0.45\textwidth}{!}{ 
\begin{tabular}{l|c|c|c} 
\toprule
\textbf{Parameter} & \textbf{Liver} & \textbf{Kidney} & \textbf{Prostate} \\
\midrule
Approximate size & 11--25 cm & 10--13 cm & 3--4 cm \\

No. of forces & 1--3 & 1--3 & 1--2 \\

Max force & 1.5 N & 1.0 N & 0.3 N \\

Force radius & 10--150 mm & 8--50 mm & 2--5 mm \\

Fixed BC radius & 25--75 mm & 15--40 mm & 5--25 mm \\

Max perturbation & 8 mm & 5 mm & 1 mm \\

No. of instances & 300 & 360 & 160 \\
\bottomrule
\end{tabular}
} 
\label{tab:organ-simulation-parameters}
\vspace{-10pt}
\end{table}

\subsubsection{Phantom dataset}
The Image-to-Physical Liver Registration Sparse Data Challenge \cite{brewer2019image} provides a publicly available phantom dataset and analysis tools for benchmarking non-rigid image-to-patient registration algorithms. This dataset features a 3D-printed preoperative liver phantom modeled from a real patient's CT image volume. Intraoperative surfaces were captured under four different deformation states, with varying surface coverage to simulate operating room conditions. The intraoperative surfaces are provided as sparse, unstructured point clouds. Approximately 75\% of the point clouds were collected using an optically tracked stylus by physically swabbing the surface of the liver phantom, while the remaining 25\% were obtained through non-contact digitization. This design allows the evaluation of algorithm performance under different point cloud sampling characteristics.

\begin{figure}[!t]
\centerline{\includegraphics[width=\columnwidth]{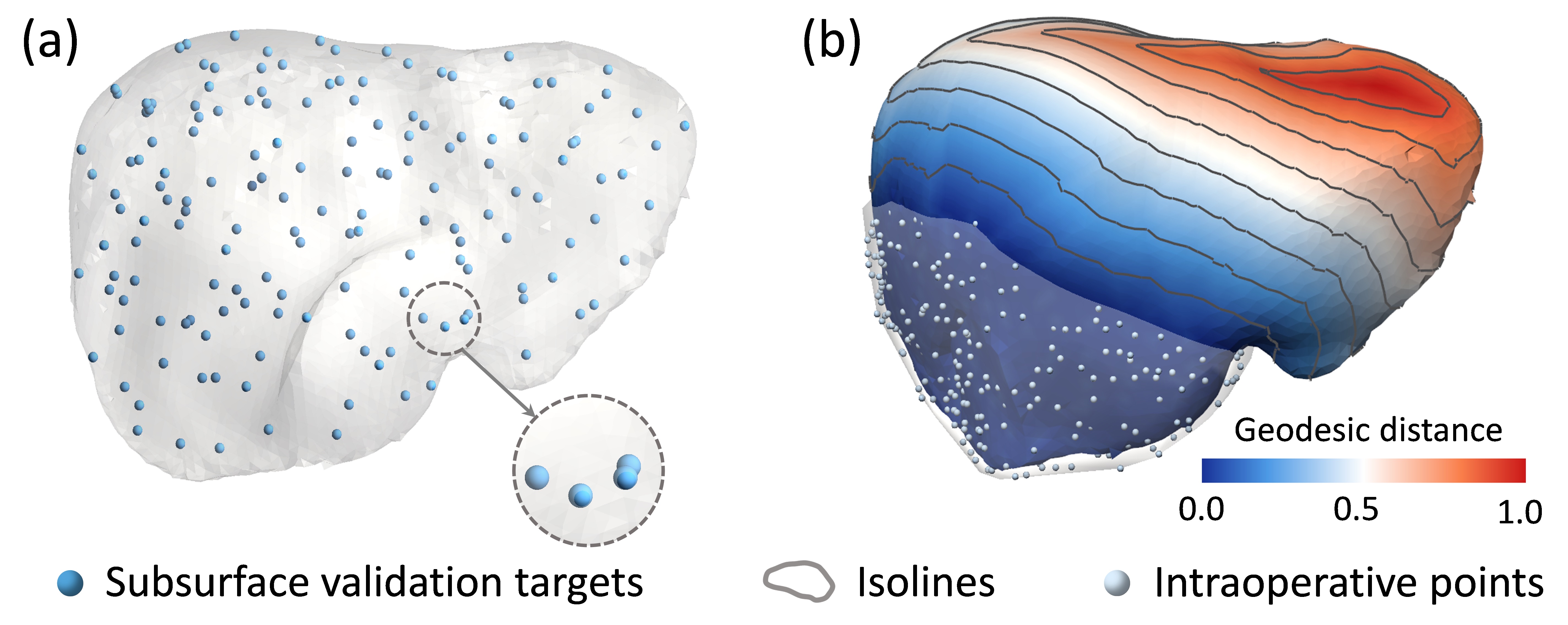}}
\caption{(a) Distribution of volumetric validation targets throughout the liver volume. A magnified view highlights the local variation in depth relative to the liver surface. (b) Visualization of normalized geodesic distance from intraoperatively observed points (semi-transparent region) to unobserved regions. Color gradient encodes relative distance, and isolines delineate equal-distance contours.}
\label{fig:liver_model}
\vspace{-10pt}
\end{figure}

For volume-level evaluation, the phantom contains 159 stainless steel beads embedded within its interior, serving as validation targets. Their spatial distribution is shown in Fig.~\ref{fig:liver_model}(a). The beads were localized through CT scans for each deformation state and paired with the corresponding deformed surface. The evaluated algorithms were tasked with estimating the volumetric deformation field of the preoperative organ based on sparse intraoperative point clouds.

To ensure compatibility with the provided accuracy analysis tools, we used the estimated liver surface displacements as boundary conditions (i.e., \( \mathbf{u}_c \equiv \mathbf{u} \) in Eq.~\eqref{eq:forward_solution}) and solved the forward problem on the officially provided volumetric mesh to obtain the full deformation field \( \mathbf{u}_{\Omega} \). The resulting predictions were then evaluated by computing the target registration errors at the 159 annotated bead locations.

\subsubsection{Human in-vivo dataset}
To evaluate the effectiveness of our method for navigated laparoscopy, we conducted both qualitative and quantitative assessments on laparoscopic liver resection scenarios. The preoperative liver model was reconstructed from CT scans, and paired laparoscopic images were captured using a monocular laparoscope. The intraoperative liver surface was extracted through manually annotated liver masks and reconstructed using the Depth Anything model~\cite{depthanything} with the calibrated camera parameters. Scale recovery and rigid alignment were performed by selecting corresponding point pairs between the preoperative and intraoperative liver models. The patient cases involved in this study are from the Self-P2IR dataset \cite{11016089}.

\subsection{Evaluation Metrics}

The mean target registration error (\(\text{TRE}\)) is adopted as the primary evaluation metric for both synthetic and phantom datasets. It measures the average Euclidean distance between the predicted and ground truth positions of target points:
\begin{equation}
    \text{TRE} = \frac{1}{N} \sum_{i=1}^N \| \mathbf{t}_{i, \text{GT}} - \mathbf{t}_{i, \text{est}} \|_2,
\end{equation}
where \( \mathbf{t}_{i, \text{GT}} \) represents the ground truth position of target \( i \), \( \mathbf{t}_{i, \text{est}} \) is the estimated position of the same target, and \( N \) is the total number of evaluated target points.

For the synthetic dataset, TRE is calculated for all nodes on the surface mesh \( \partial \Omega \), leveraging the availability of ground truth positions for each node under the applied forces and boundary conditions. In this case, the \(\text{TRE}\) metric assesses surface-to-surface registration performance.

For the phantom dataset, \(\text{TRE}\) is computed for 159 stainless steel beads embedded inside the liver phantom. The ground truth positions \( \mathbf{t}_{i, \text{GT}} \) are obtained via CT imaging, while the estimated positions \( \mathbf{t}_{i, \text{est}} \) are derived by propagating the predicted surface displacements from the surface mesh \( \partial \Omega \) to the volumetric mesh \( \Omega \), as detailed in Sec. II-C.

For the human in-vivo dataset, where ground truth 3D deformation is not directly observable, we compute a one-sided Chamfer Distance (CD) from the intraoperatively reconstructed liver surface \( \mathcal{Y} \) to the deformed preoperative model \( \mathcal{X} + \mathcal{U} \). This metric evaluates surface alignment accuracy in visible regions and serves as a quantitative indicator of registration fidelity in real surgical scenarios. The one-sided CD is defined as:
\begin{equation}
    \text{CD}(\mathcal{Y}, \mathcal{X} + \mathcal{U}) = \frac{1}{|\mathcal{Y}|} \sum_{\mathbf{y} \in \mathcal{Y}} \min_{\mathbf{x} \in \mathcal{X}} \left\| \mathbf{y} - (\mathbf{x} + \mathbf{u}) \right\|_2^2,
\end{equation}
where \( \mathcal{X} \) denotes the preoperative liver surface, \( \mathcal{U} \) the predicted deformation field, and \( \mathcal{Y} \) the intraoperatively reconstructed partial surface.

\subsection{Implementation Details}
All experiments were conducted on a single Nvidia RTX 3090 GPU, running on a 3.60 GHz Intel\textsuperscript{®} Xeon\textsuperscript{®} W-2223 CPU with 32 GB of RAM. The implementation details for each component of the framework are as follows: \\
\textbf{BiomPINN:} The neural network follows a four-level pyramid structure, where each level consists of an MLP with a depth of 3 layers and a width of 64 neurons per layer. The Adam optimizer is used with the initial learning rate of 0.01. The balance factors \( \lambda_1 \) and \( \lambda_2 \) for the regularization terms were tuned individually for each organ within the range \( [10^{-6}, 10^{-3}] \), in order to account for organ-specific differences in deformation scale and tissue response. \\
\textbf{PBM:} The biomechanical model is filled with a linearly elastic material, constructed using the FE-meshing tool TetGen \cite{hang2015tetgen}. The material properties of the model are defined by a Young's modulus of 5 kPa and a Poisson's ratio of 0.35. The iterative conjugate gradient solver, \(\texttt{sparse.linalg.cg}\), from the \texttt{scipy} library is employed to solve the sparse linear system in Eq.~\eqref{eq:pbm_linear_system}, with the convergence tolerance set to \(10^{-5}\). \\
\textbf{Human-Computer Interaction System:} The interactive system is developed on the Microsoft HoloLens 2 (HL2) platform, leveraging the open-source Mixed Reality Toolkit. Due to the computational limitations of the head-mounted device, the HL2 is dedicated solely to rendering visualizations (e.g., preoperative and intraoperative models) and capturing surgeon input through line-drawing prompts. The optimization for deformation modeling based on the received prompts is performed remotely on a server. Communication between the server and the HL2 is handled via asynchronous TCP socket programming. For all experiments involving human-computer interaction, the annotation process was performed by two users: one male computer science student and one female biomedical engineering student.

\section{Results}
\subsection{Synthetic Dataset Validation}
We begin by evaluating the ability of BiomPINN and BiomPINN-PBMs to approximate deformation fields generated by an FE-based approach. This analysis also examines the contribution of biomechanical regularization to prediction accuracy. GMM-FEM~\cite{khallaghi2015biomechanically} is chosen as a baseline for comparison, as it shares the same correspondence scheme with BiomPINN and BiomPINN-PBMs, enabling a controlled assessment of the added value from the FE-based regularization.

\begin{figure*}[!t]
\centering
\includegraphics[width=0.98\textwidth]{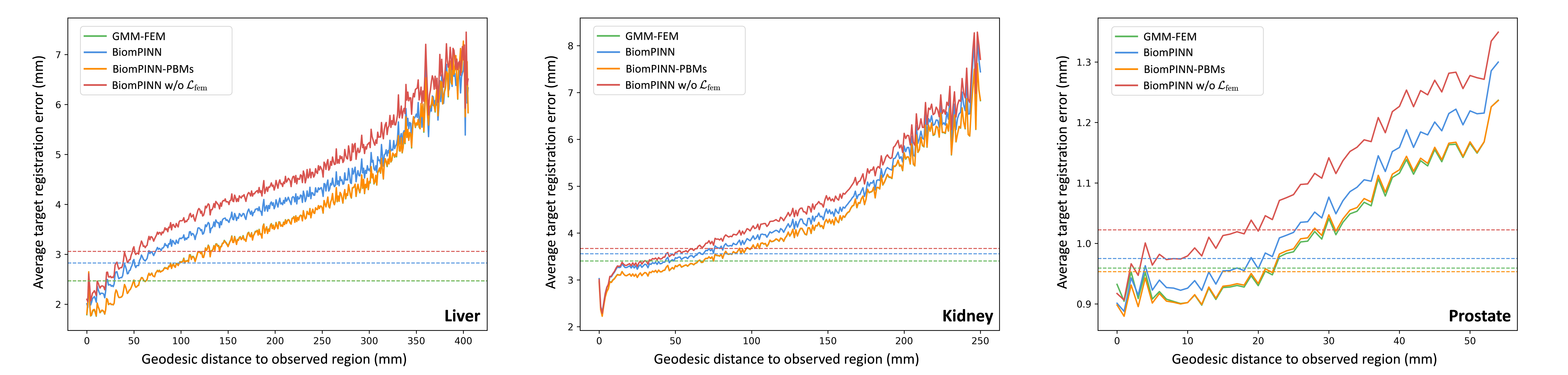}
\caption{Performance comparison on the synthetic dataset across liver, kidney, and prostate. 
The plots show how average target registration error varies with geodesic distance from observed regions, reflecting each method’s ability to propagate deformation to unobserved areas.}
\label{fig:synthetic}
\vspace{-10pt}
\end{figure*}

Fig.~\ref{fig:synthetic} illustrates how prediction accuracy varies with geodesic distance from observed intraoperative regions, evaluated across liver, kidney, and prostate. Regions with a geodesic distance of 0 correspond to areas directly observed during surgery, where the deformation error reflects local alignment accuracy. As the geodesic distance increases, prediction errors reveal each method’s ability to propagate deformation to unobserved regions. The geodesic distance mapping on the organ surface is visualized in Fig.~\ref{fig:liver_model}(b).

\begin{table}[t]
\centering
\caption{TRE (Mean ± Std) [mm] and runtime [s] for different organs on the Synthetic Dataset}
\resizebox{0.45\textwidth}{!}{
\begin{tabular}{c|c|c|c}
\toprule
\textbf{Organ} & \textbf{Methods} & \textbf{TRE (\(\mu \pm \sigma\))} & \textbf{Runtime} \\
\midrule
\multirow{4}{*}{Liver} 
& GMM-FEM \cite{khallaghi2015biomechanically} & 2.47 ± 1.16 & 4.28 \\
& BiomPINN & 2.82 ± 1.30 & 0.82 \\
& BiomPINN-PBMs & \textbf{2.47 ± 1.15} & 1.65 \\
& BiomPINN (w/o \(\mathcal{L}_{\text{fem}}\)) & 3.05 ± 1.31 & 0.23 \\
\midrule
\multirow{4}{*}{Kidney} 
& GMM-FEM \cite{khallaghi2015biomechanically} & \textbf{3.40 ± 2.40} & 3.18 \\
& BiomPINN & 3.56 ± 2.46 & 0.57 \\
& BiomPINN-PBMs & \textbf{3.40 ± 2.40} & 1.25 \\
& BiomPINN (w/o \(\mathcal{L}_{\text{fem}}\)) & 3.67 ± 2.37 & 0.23 \\
\midrule
\multirow{4}{*}{Prostate} 
& GMM-FEM \cite{khallaghi2015biomechanically} & 0.96 ± 0.74 & 3.38 \\
& BiomPINN & 0.98 ± 0.78 & 0.58 \\
& BiomPINN-PBMs & \textbf{0.95 ± 0.75} & 1.50 \\
& BiomPINN (w/o \(\mathcal{L}_{\text{fem}}\)) & 1.02 ± 0.75 & 0.23 \\
\bottomrule
\end{tabular}
}
\label{tab:synthetic_results}
\vspace{-10pt}
\end{table}

Across all organs, prediction error increases with geodesic distance. Notably, all four methods exhibit similar performance in regions with zero geodesic distance, indicating comparable alignment accuracy in directly observed areas. However, in unobserved regions, models incorporating biomechanical constraints (BiomPINN and BiomPINN-PBMs) achieve lower errors than the unconstrained variant, demonstrating their ability to propagate deformation in a physically plausible manner beyond the observed surface.

On the liver dataset, BiomPINN-PBMs achieves a mean TRE statistically equivalent to GMM-FEM (\( p = 0.21 \), paired \( t \)-test), while reducing runtime from 4.28 to 1.65~seconds per case. A similar trend is observed for the kidney dataset, where the TRE difference is not statistically significant (\( p = 0.36 \)), and BiomPINN-PBMs again reduces runtime from 3.18 to 1.25~seconds. Remarkably, on the prostate dataset, BiomPINN-PBMs outperforms GMM-FEM with a statistically significant improvement in TRE (\( p = 0.0001 \), \( t = 4.15 \)), indicating that the observed performance gain is unlikely due to random variation. These results highlight the robustness of BiomPINN-PBMs across varying anatomical structures and deformation scales. All reported runtimes represent the total time to compute one deformation field per case. For BiomPINN-PBMs, the runtime includes the full per-instance optimization process and the solution of the biomechanical system. In contrast, the runtime of GMM-FEM only reflects the time required to solve the linear FE system. This setup allows the runtimes to be directly comparable across methods. Detailed quantitative results are summarized in Table~\ref{tab:synthetic_results}.

\begin{figure}[!t]
\centerline{\includegraphics[width=0.96\columnwidth]{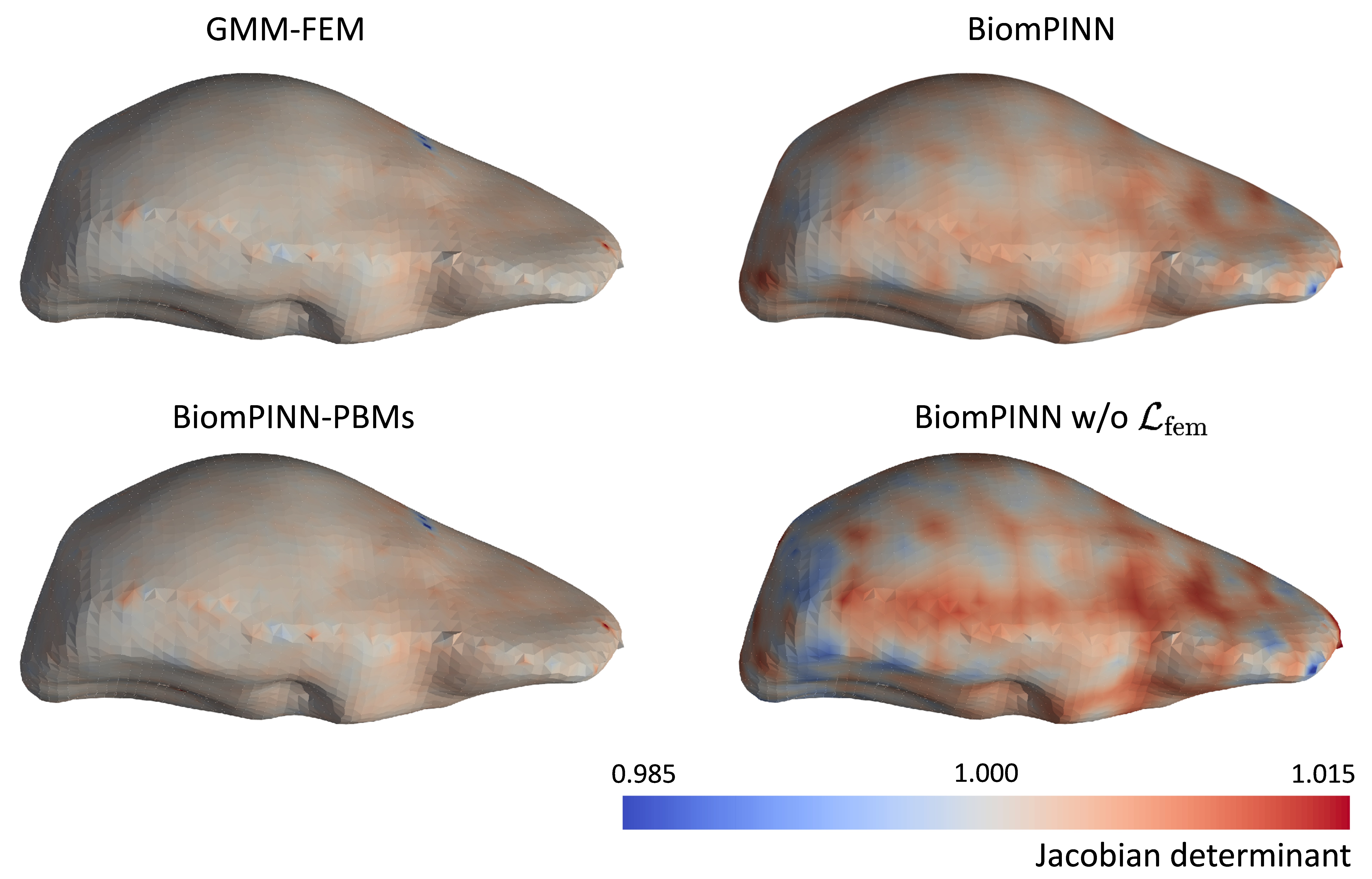}}
\caption{Field consistency assessment: Jacobian determinant heatmap.}
\label{fig:consistency}
\vspace{-5pt}
\end{figure}

Furthermore, we evaluate field consistency using the Jacobian determinant \(|J|\) as a metric, defined as \( |J| = |\nabla u + I| \), where \(\nabla u\) is the displacement gradient tensor, \(I\) is the identity matrix, and \(|\cdot|\) denotes the matrix determinant. This metric measures local volume change and is expected to approach unity for nearly incompressible soft tissues. The heatmap of Jacobian determinant values across the volumetric tetrahedral mesh \(\Omega\), evaluated on the liver model, is shown in Fig.~\ref{fig:consistency}. BiomPINN-PBMs exhibits a distribution of Jacobian determinants closely aligned with that of GMM-FEM. BiomPINN shows moderate consistency, with slightly higher variability. In contrast, removing \(\mathcal{L}_{\text{fem}}\) leads to scattered and irregular Jacobian values, indicating inconsistencies in the deformation field. These results underscore the critical role of biomechanical regularization in ensuring physically plausible deformations.

\subsection{Phantom Dataset Validation}
Experiments on the phantom dataset~\cite{brewer2019image} were conducted to evaluate the performance of our framework under two key factors: (1) different point cloud sampling characteristics and (2) varying intraoperative visibility conditions. To ensure a fair benchmark, the deformation patterns and the distribution of internal validation targets were blinded throughout the evaluation, following the standard protocol for this dataset.

\begin{table}[t]
\centering
\caption{TRE (Mean ± Std) [mm] under different point cloud sampling strategies on the phantom dataset.}
\resizebox{0.45\textwidth}{!}{
\begin{tabular}{c|c|c}
\toprule
\textbf{Methods} & \textbf{Swabbed Surface} & \textbf{Digitized Surface} \\
\midrule
Rigid Alignment & 5.70 ± 1.96 & 5.06 ± 1.10 \\ 
Ours (w/o prompt) & 3.45 ± 0.74 & 3.38 ± 0.65 \\ 
Ours (w/ prompt) & \textbf{2.80 ± 0.70} & \textbf{2.74 ± 0.63} \\ 
\bottomrule
\end{tabular}
}
\label{tab:phantom_results1}
\vspace{-10pt}
\end{table}

\begin{figure}[!t]
\centerline{\includegraphics[width=0.96\columnwidth]{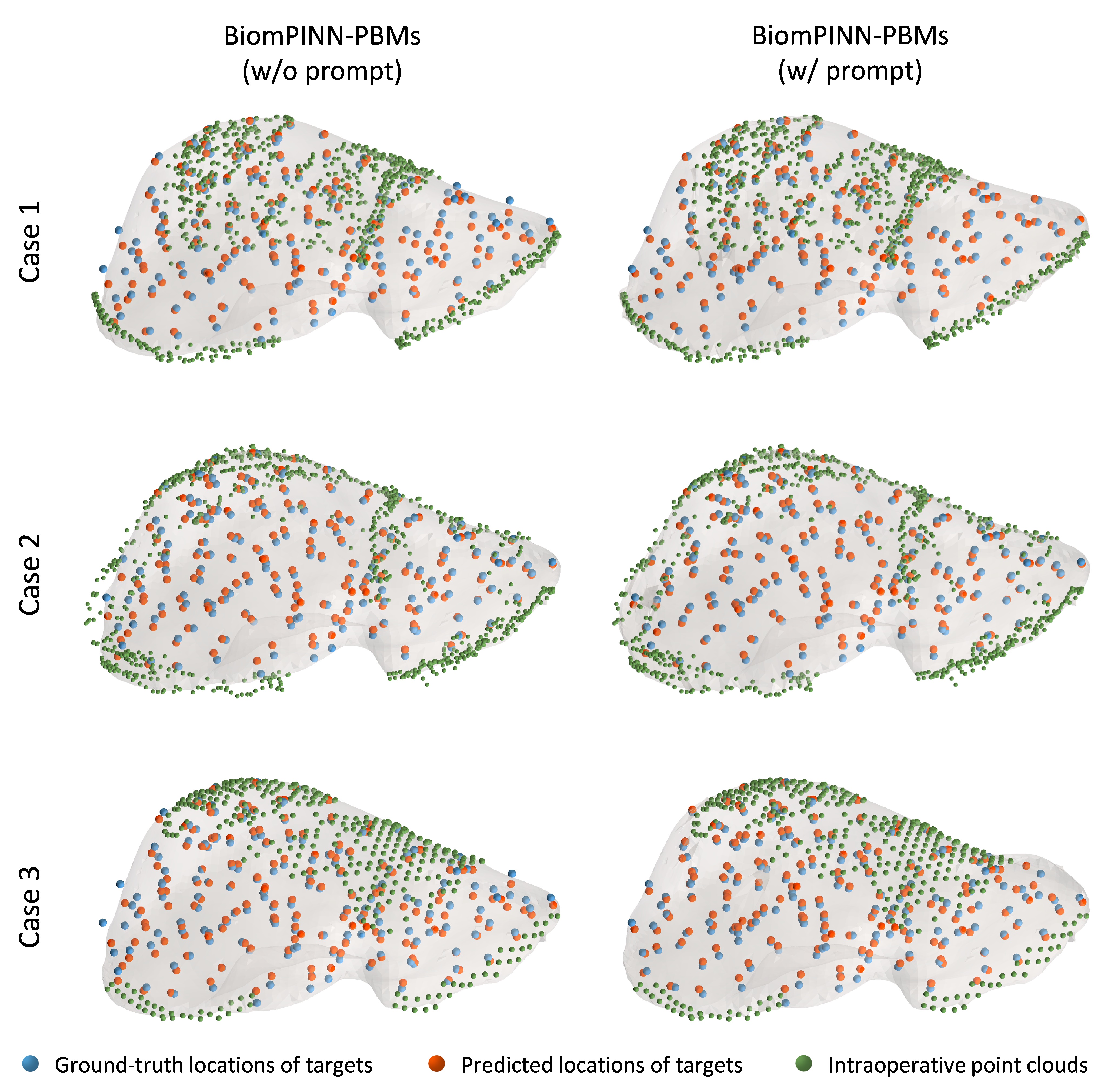}}
\caption{Phantom dataset qualitative results. Comparison between fully automated deformation (left) and surgeon-refined deformation after prompts (right) across different cases. Green: intraoperative point clouds; Blue: ground-truth target locations; Orange: predicted target locations. Cases 1 and 2: physically swabbed point clouds. Case 3: non-contact digitized point cloud.}
\vspace{-10pt}
\label{fig:phantom}
\end{figure}

First, we assessed the impact of different intraoperative point cloud acquisition strategies. In this dataset, approximately 75\% of the intraoperative surfaces were collected through physical swabbing of the liver phantom using an optically tracked stylus, resulting in locally varying point densities due to manual acquisition variability. The remaining 25\% were acquired via non-contact digitization, yielding more uniformly distributed point clouds. Figure~\ref{fig:phantom} illustrates examples of the two sampling approaches: Cases 1 and 2 correspond to manually swabbed surfaces, while Case 3 corresponds to a uniformly sampled surface obtained through non-contact digitization. Table~\ref{tab:phantom_results1} summarizes the quantitative results. Although variations in point cloud sampling introduced differences in initial rigid alignment errors, which were estimated using a RANSAC-based registration method, their effect on final deformation modeling accuracy was relatively minor. Incorporating human prompts further improved the \(\text{TRE}\) by more than 0.6~mm across both acquisition strategies.

\begin{table*}[t]
\centering
\caption{TRE (Mean ± Std) [mm] under different surface coverage conditions on the phantom dataset.}
\resizebox{0.85\linewidth}{!}{ 
\begin{tabular}{c|c|ccc|c} 
\toprule
\multirow{2}{*}{\textbf{Methods}} & \multirow{2}{*}{\textbf{Type}} & \multicolumn{3}{c|}{\textbf{Surface Coverage}} & \multirow{2}{*}{\textbf{All}} \\ 
\cmidrule(lr){3-5}
                                   &                                & 20-28\%        & 28-36\%        & 36-44\%        &                              \\ 
\midrule
V2S-Net \cite{pfeiffer2020non} (2020)    & \multirow{2}{*}{Deep Learning} & 7.79 ± 4.86    & 6.79 ± 4.63    & 8.24 ± 6.34    & 7.55 ± 5.28                  \\
Jia et al. \cite{jia2021improving} (2021) &                                & 4.22 ± 0.96    & 4.57 ± 2.79    & 4.02 ± 1.31    & 4.29 ± 1.93                  \\ 
\midrule
Heiselman et al. \cite{heiselman2020intraoperative} (2020) & \multirow{5}{*}{Biomechanical Model} & 3.27 ± 1.07    & 3.00 ± 0.67    & 2.99 ± 0.78    & 3.08 ± 0.85                  \\
Mestdagh et al. \cite{mestdagh2022optimal} (2022)  &                                & 3.54 ± 1.11    & 3.27 ± 0.85    & 3.13 ± 0.82    & 3.31 ± 1.86                  \\
Ringel et al. \cite{ringel2023comparing} (2023)  &                                & 4.56 ± 0.88    & 4.62 ± 0.99    & 4.62 ± 0.98    & 4.60 ± 0.95                  \\
Yang et al. \cite{yang2024boundary} (2024)   &     & 3.05 ± 0.75 & 2.94 ± 0.66 & 2.78 ± 0.68 & 2.93 ± 0.68 \\
Heiselman et al. \cite{heiselman2024optimal} (2024) &                                & 5.06 ± 0.92    & 4.91 ± 1.11    & 4.61 ± 1.14    & 4.86 ± 1.07  \\
\midrule
BiomPINN-PBMs (w/o prompt) & \multirow{2}{*}{Hybrid Method} & 3.56 ± 0.75    & 3.41 ± 0.67    & 3.31 ± 0.73    & 3.42 ± 0.72                  \\
BiomPINN-PBMs (w/ prompt) &                                & \textbf{2.91 ± 0.75}     & \textbf{2.79 ± 0.64}     & \textbf{2.64 ± 0.65}   & \textbf{2.78 ± 0.68}                 \\ 

\bottomrule
\end{tabular}
}
\label{tab:phantom_results}
\vspace{-5pt}
\end{table*}

Second, we evaluated performance under different levels of intraoperative surface visibility, defined by the percentage of the liver surface observed in the intraoperative point cloud (20–28\%, 28–36\%, and 36–44\%). We compared our method against all prior approaches evaluated on this dataset, and the results of the competing methods are consistent with those reported in previous studies~\cite{heiselman2024image, yang2024boundary}. Table~\ref{tab:phantom_results} presents the quantitative comparison. In the fully automated setting (without human prompts), BiomPINN-PBMs achieved a \(\text{TRE}\) of 3.42~mm across all cases. When surgeon prompts were incorporated, the \(\text{TRE}\) further improved to 2.78~mm, surpassing all existing methods. These results highlight the effectiveness of our interactive deformation modeling framework in enabling surgeon-driven refinements.

Qualitative examples are shown in Fig.~\ref{fig:phantom}, demonstrating improvements in alignment between preoperative models and intraoperative point clouds with human prompts. Specifically: Case 1 shows correction of mismatches along the liver's inferior ridges; Case 2 demonstrates improved alignment in the right lobe region; and Case 3 highlights local correction of mismatches in the right inferior ridge with propagated improvements to unobserved areas, such as the right lobe.

\subsection{Human In-Vivo Dataset Validation}

\begin{figure*}[!t]
\centering
\includegraphics[width=0.95\textwidth]{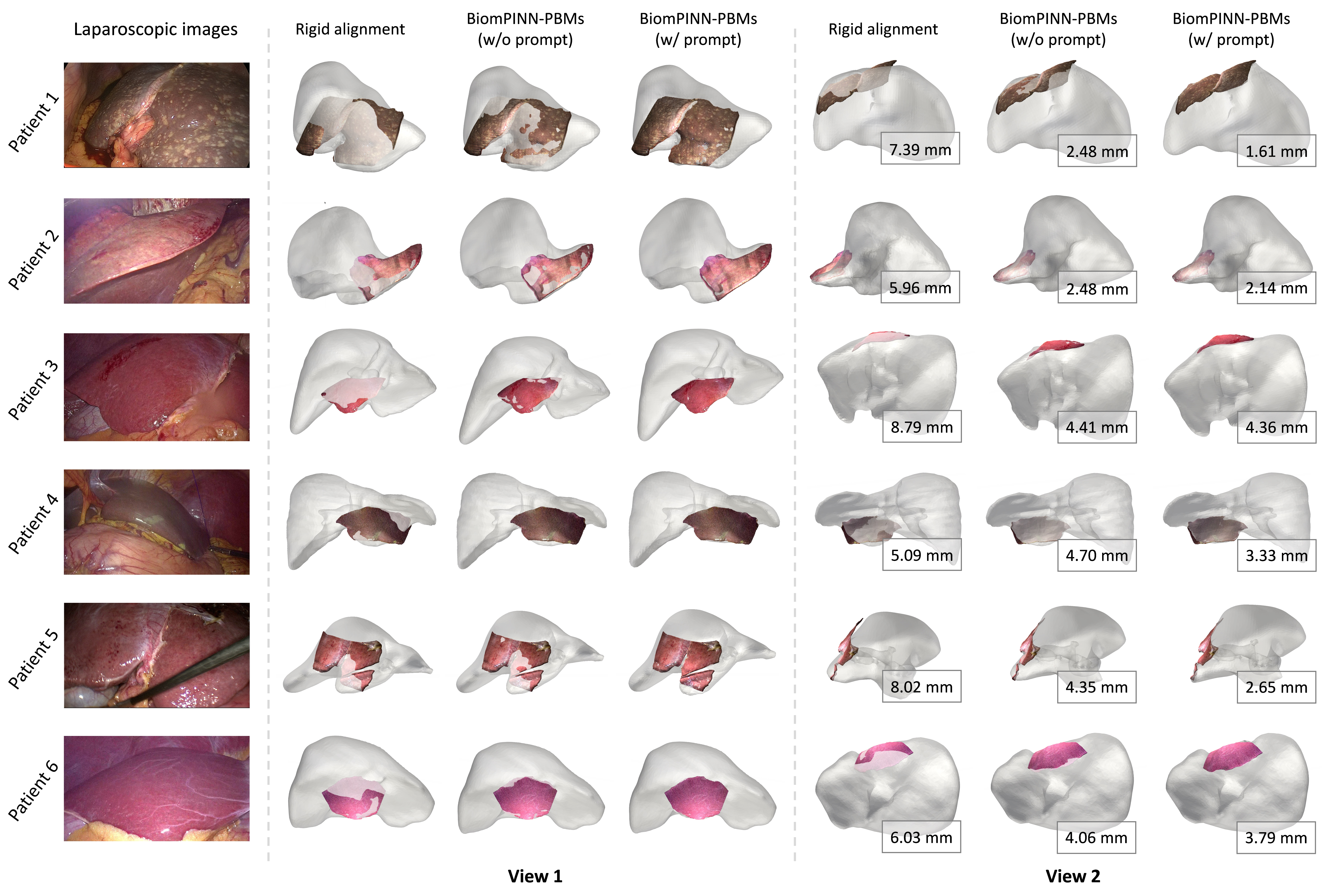}
\caption{Qualitative and quantitative results on laparoscopic liver surgery cases. Each row corresponds to one patient, showing two different views (View 1 and View 2) of the surface alignment outcomes. For each patient, we compare rigid alignment, BiomPINN-PBMs without prompts, and BiomPINN-PBMs with prompts. Overlaid values denote one-sided CD (mm) between the intraoperative surface and the deformed preoperative model. Lower values indicate more accurate alignment.}
\label{fig:in_vivo_results}
\vspace{-10pt}
\end{figure*}

We first evaluate surface alignment accuracy using one-sided CD from the deformed preoperative model to the intraoperatively reconstructed liver surface. As shown in Fig.~\ref{fig:in_vivo_results}, BiomPINN-PBMs consistently reduces registration error compared to rigid alignment. Further reductions are observed when surgeon prompts are incorporated, highlighting the utility of interactive guidance in refining deformation estimates.

Qualitative results from individual human in-vivo cases further illustrate the strengths and limitations of the automated framework. In certain cases, the automatic method yields anatomically consistent alignment without requiring user input. For instance, in Patient 2, BiomPINN-PBMs effectively captures the deformation of the left liver lobe in accordance with intraoperative observations. A comparable result is observed in Patient 6, where the model achieves accurate surface correspondence in the absence of human intervention.

In more challenging scenarios, particularly those involving topological alterations or ambiguous structures, surgeon input plays a critical role. In Patient 1, resection of the falciform ligament leads to substantial separation between the liver lobes, which cannot be adequately resolved by the automatic method alone. Following surgeon delineation of the interlobar boundary, the resulting deformation field more accurately reflects the anatomical configuration. In Patient 3, although surface alignment is reasonably close, the left-right lobe boundary is misregistered. Surgeon correction resolves this inconsistency, even though the change in CD is minimal. In Patient 4, targeted adjustment near the lower ridge of the left lobe further enhances local anatomical accuracy. For Patient 5, the model successfully aligns the right lobe automatically, whereas the left lobe requires user guidance.

These findings suggest that while the automatic framework is capable of producing reliable deformation results in terms of quantitative metrics, surgeon involvement remains essential for ensuring anatomical plausibility in complex or altered surgical anatomies.

\subsection{End-to-End Human-in-the-Loop Time Analysis}
We performed a detailed analysis of the time required for a single surgeon-prompt interaction cycle, measuring from the initiation of user annotation to the point when the updated deformation result is rendered for surgeon evaluation. The reported timings exclude cognitive decision-making and reflect purely mechanical annotation and system processing latency.

Among five repeated tests, the longest end-to-end time for the human-in-the-loop correction process was recorded as~20.64 seconds, detailed as follows:
\begin{itemize}
    \item User annotation and prompt submission (17.80~s total): 
    \begin{itemize}
        \item Drawing the line-based prompts: 13.70~s.
        \item Submitting annotations to the remote server: 4.10~s.
    \end{itemize}
    \item Server-side processing (2.72~s total):
    \begin{itemize}
        \item ICP for correspondence matrix refinement: 1.05~s.
        \item Data preprocessing (network initialization, CPU-GPU data transfer, etc.): 0.46~s.
        \item Computing refined deformation field: 1.21~s.
    \end{itemize}
    \item Wireless transmission (0.12~s total):
    \begin{itemize}
        \item Transmitting updated deformation results back to surgeon within a local area network: 0.12~s.
    \end{itemize}
\end{itemize}

The results indicate that the annotation phase constitutes the primary bottleneck, accounting for approximately 87\% of the total cycle time. Therefore, future improvements should prioritize optimizing the user annotation process, potentially through enhanced interaction modalities (e.g., touch-based annotation on surgical displays), semi-automated prompt generation, or more efficient data submission mechanisms, in order to further reduce latency and improve intraoperative usability.

\section{DISCUSSION}
Accurate intraoperative deformation modeling is a critical prerequisite for achieving reliable AR-guided surgical navigation. Despite significant progress, fully automated pipelines remain insufficient to robustly handle the complexities of real-world clinical scenarios. To address this limitation, we introduce a human-in-the-loop deformation modeling framework that incorporates surgeon-provided annotations (prompts) to correct misaligned anatomical regions. By integrating expert knowledge into the computational modeling process, our approach ensures anatomically plausible and reliable deformation predictions, even in challenging surgical contexts. This interactive correction mechanism not only enhances the robustness of AR navigation but also improves the clinical feasibility and acceptance of computer-assisted surgical interventions.

In parallel, to meet clinical demands for rapid response times, we propose BiomPINN-PBMs, a data-driven biomechanics algorithm designed to enhance computational efficiency. This framework integrates a finite-element-informed neural optimization scheme (BiomPINN) into the patient-specific biomechanical modeling process, enabling direct prediction of the regularization weight required for deformation estimation. By avoiding iterative optimization that requires repeated solutions of large-scale linear systems, our method retains the accuracy of FE-based approaches while enabling a one-pass solution to the volumetric deformation problem, thereby reducing computational time.

To better contextualize our contributions and identify areas for further improvement, we analyze the strengths, limitations, and potential future directions of the proposed framework from three key perspectives: deformation modeling algorithms, clinical applicability, and human-computer interaction.

\subsection{Deformation Modeling Algorithm}
First, we adopt a mesh-based representation to model organ geometry and deformation. Compared to voxel grids, meshes inherently avoid resolution limitations and voxelization artifacts, preserving fine anatomical details critical for surgical navigation. Moreover, unlike point cloud or implicit neural field representations, meshes explicitly encode topological connectivity among nodes, which is essential for incorporating biomechanical regularization. The stiffness matrix derived from the FE formulation naturally links local deformations across neighboring elements, enabling physically plausible deformation fields that respect tissue mechanics—an important property for anatomically realistic intraoperative modeling.

Despite the structural advantages of mesh-based representations, the accuracy of deformation estimation remains highly dependent on the quality of intraoperative point cloud acquisition. In our framework, surface correspondences established between the preoperative mesh and intraoperative point clouds drive deformation modeling. Thus, the fidelity of the reconstructed intraoperative surface directly influences performance. Although phantom experiments demonstrate robustness to variations in point cloud density and sampling distribution, an implicit assumption is that the point cloud accurately captures the true organ surface without significant contamination from surrounding tissues or instruments. In real surgical environments, achieving such clean segmentation is challenging, making reliable organ surface extraction a critical prerequisite for robust modeling.

Furthermore, the type of intraoperative surrogate information used to constrain deformation plays a crucial role in determining modeling accuracy. While surface-based signals are practical and readily obtainable through visual modalities, they inherently lack information about deeper volumetric structures, limiting the ability to accurately capture internal deformations. Future directions could include integrating richer intraoperative signals—such as vascular features extracted from ultrasound imaging—or incorporating physiological measurements like respiratory volume changes for modeling highly dynamic organs such as the lungs.

Finally, our current framework is designed to support on-demand deformation correction during critical surgical moments, such as verifying the location of subsurface vessels or tumors to determine surgical pathways. However, future navigation systems should aim for continuous intraoperative guidance, placing greater demands on real-time performance. Leveraging learning-based strategies to develop high-performance preconditioners presents a promising avenue for accelerating the biomechanical system solution phase \cite{li2023learning}. Integrating such approaches into our framework offers the potential to enable reliable and real-time deformation modeling for adaptive surgical navigation.

\subsection{Clinical Applicability}
Beyond algorithmic performance, the clinical applicability of deformation modeling frameworks ultimately determines their translational value. A key strength of our approach lies in its interactive human-in-the-loop mechanism, which enables surgeons to enhance navigation accuracy by indicating anatomical correspondences based on their expertise. According to surgeon feedback, during certain critical moments, accurate localization of vascular structures is far more valuable than having a continuously updated yet less reliable model. For example, when determining the surgical approach to the hepatic hilum, surgeons emphasized that if they are given the opportunity to correctly localize vessels before making any irreversible action, even a two-minute surgeon-guided correction to refine anatomical alignment is considered worthwhile. The ability to obtain accurate guidance during such key decision points justifies the additional effort.

Another potential clinical advantage of our framework lies in its ability to support the seamless integration of preoperative surgical planning—such as resection margins, ablation zones, or planned surgical paths—onto the intraoperative anatomy through deformation modeling. This is made possible by our mesh-based representation, which preserves anatomical topology even after deformation. Such capability not only allows surgeons to visualize planned surgical trajectories during the procedure, but also paves the way for real-time surgical plan updates based on intraoperative changes, representing a critical step toward adaptive and personalized surgery. Furthermore, preoperatively defined resection or ablation volumes may also serve as additional constraints during deformation modeling, further improving alignment accuracy and enhancing the clinical utility of the guidance system.

In addition to intraoperative navigation, the utility of interactive deformation modeling extends to broader clinical scenarios requiring accurate anatomical alignment. For instance, in postoperative liver resection assessment, surgeons often seek to evaluate deviations between the actual resection plane and preoperatively planned margins, necessitating non-rigid registration between pre- and postoperative anatomies. Current automated methods often struggle to achieve precise alignment, leaving surgeons without practical means to intervene. Our framework provides a direct mechanism for surgeon-guided corrections in regions of mismatch, thereby improving the reliability and clinical utility of registration outcomes.

Nonetheless, we recognize practical challenges that must be addressed for seamless clinical integration. At present, surgeon prompts are provided through a head-mounted AR device, which, while effective for proof-of-concept validation, introduces additional equipment requirements during surgery. To reduce workflow disruption, we are actively developing alternative interfaces that allow prompt annotations directly on surgical displays (e.g., touchscreens), facilitating more natural integration into existing operating room environments.

\subsection{Human-Computer Interaction}
From a human-computer interaction perspective, our current line-drawing annotation approach offers surgeons an intuitive and direct means of manipulating deformation predictions. Nonetheless, a critical factor not extensively explored in this study is the influence of surgeon annotation accuracy on the overall deformation modeling outcomes. Here, annotation accuracy refers to how precisely a surgeon can spatially delineate anatomical misalignments relative to the intended targets within the AR environment.

To preliminarily assess the achievable spatial annotation accuracy with mid-air AR interactions, we conducted a small-scale user study. Participants familiar with AR headset operation were instructed to reproduce predefined spatial paths by drawing the shape of a reference line as precisely as possible in mid-air. Results showed a bidirectional Chamfer Distance annotation error of approximately 5–7 mm (mean 6.13 mm) in absolute physical space. Nevertheless, given the flexibility to arbitrarily scale virtual organ models during annotation, the practical impact of this absolute error is context-dependent: at larger magnifications, such discrepancies may be negligible, whereas at smaller scales, they could substantially affect deformation modeling accuracy.

Quantifying how annotation errors propagate into final deformation metrics remains an open challenge and warrants further investigation. Importantly, this consideration highlights the need for immediate visual feedback during the annotation process. For instance, real-time visualization of point correspondences derived from user prompts would allow surgeons to promptly assess and, if necessary, adjust their annotations to improve correction fidelity. Integrating such feedback mechanisms could significantly enhance both the usability and robustness of the human-in-the-loop interaction.

\section{CONCLUSION}
This study contributes to addressing intraoperative organ deformation, a critical step toward achieving reliable AR-guided surgical navigation. We propose a data-driven biomechanical model that achieves accuracy comparable to traditional FE-based methods while significantly improving computational efficiency. Furthermore, we introduce a human-in-the-loop interaction framework that enables surgeons to actively refine deformation predictions during complex surgical scenarios, enhancing the adaptability and reliability of AR navigation. Extensive qualitative and quantitative experiments demonstrate the effectiveness of our interactive approach, highlighting its potential for clinical translation. Future work will focus on expanding surgeon interaction modalities, such as touch-based annotations and voice commands, to improve flexibility and clinical integration. We also plan to incorporate richer intraoperative data—such as vascular features extracted from ultrasound imaging—to better capture internal deformations and further enhance volumetric modeling fidelity.

\bibliographystyle{IEEEtran}  
\bibliography{reference}     

\end{document}